# EEGHyena2: Fixation-Inspired Joint Lexical-Grammatical Conditional Random Field for Interpretable EEG-to-Text Decoding

**Xiaoshan Zhou[1*], Chenlan Liu[2]**

[1]School of Project Management, Faculty of Engineering, The University of Sydney, Sydney, NSW, Australia, xiaoshan.zhou@sydney.edu.au

[2]Institute of Software, Chinese Academy of Sciences, Beijing, China, liuchenlan@iscas.ac.cn

*Corresponding Author

## Highlights

- Fixation-related EEG separates word classes in group averages, not in single words.
- EEGHyena2 turns word-level EEG into lexical and grammatical evidence with two heads.
- A joint word–POS CRF scores a restricted lattice; a frozen LLM adds context scores.
- Matches published BLEU-1, raises ROUGE-1 F1 and decodes several times faster.
- Learned CRF transitions recover English syntax, e.g. determiner→adjective→noun.

## Abstract

A sentence is built from words that carry lexical meaning and grammatical roles, and brain activity during reading reflects both. Current EEG-to-text decoders, however, let a large language model (LLM) propose and order the words, so fluent output can reflect the LLM's prior more than the recorded EEG. In this study, we asked whether fixation-linked EEG carries lexical and grammatical evidence during natural reading and how a decoder can use such evidence. Fixation-related potential analyses showed that content and function words differ reliably in group-averaged EEG but only weakly in single words. We therefore propose EEGHyena2, a structured decoder in which a bidirectional encoder inspired by StripedHyena 2 feeds a semantic head that ranks a fixed lexicon and a part-of-speech (POS) head that predicts grammatical categories; a joint word–POS conditional random field (CRF) scores a restricted candidate lattice, and a frozen Llama-2 scores the candidates in context. On two natural-reading corpora, ZuCo and ROAMM, it matched the state-of-the-art method on BLEU-1 (21.4 vs 21.3), exceeded it on ROUGE-1 F1 (23.3 vs 20.3), lowered the character error rate on ROAMM (75.3% vs 84.7%) and decoded each recording about 3–10 times faster than LLM candidate search. The CRF learned transitions that mirror English syntax, and the encoder drew local context from its convolutions and distant context from attention. Decoders can thus pool weak word-level EEG evidence across the sentence through explicit lexical and grammatical structure, with the language model adjudicating among candidates rather than generating them, which lowers decoding cost and keeps decisions inspectable.

## 1. Introduction

People who lose speech and movement through amyotrophic lateral sclerosis, brainstem stroke or spinal cord injury depend on brain–computer interfaces that turn neural activity into communication without muscular output (Wolpaw et al., 2002). Invasive speech neuroprostheses now decode attempted speech into text at rates approaching conversation (Willett et al., 2023; Metzger et al., 2023), but they require implanted electrodes. Electroencephalography (EEG) is non-invasive and portable, which makes decoding language from EEG an attractive, if much harder, route. A common testbed is natural reading, in which EEG is recorded together with eye movements so that brain activity can be aligned to each fixated word (Hollenstein et al., 2018). Readers fixate a word for about 200–250 ms (Rayner, 1998), so a decoder intended for interactive use must also keep pace with its input.

Pretrained language models have made EEG-to-text decoders produce readable sentences. Early work coupled word-level EEG features to a pretrained BART decoder (Wang & Ji, 2022), and later work aligned EEG with text representations through contrastive and masked objectives (Wang et al., 2024). The current state of the art on ZuCo, LLM4EEG2Text, aligns word-level EEG with BERT semantics and lets a frozen Llama-2 propose about 300 candidate words at each position, from which the EEG representation selects one (Zheng et al., 2026). The language model thus supplies the lexical knowledge and the sentence structure that the EEG alone cannot.

However, this reliance has a cost. A capable language model shapes the probability of every word in the sentence, so a decoder can produce fluent, plausible text with little support from the EEG. Supplying the correct preceding words at test time inflates reconstruction scores, and some systems keep much of their overlap with the reference when the EEG is replaced by noise (Jo et al., 2024). More generally, language models produce fluent content that their input does not support (Ji et al., 2023). When the language model proposes the candidates, the EEG can only choose among words the model already favours, and generating those candidates word by word adds repeated large-model inference (Poddar et al., 2025; Fernandez et al., 2025; Yong et al., 2026). The challenge is to keep the linguistic knowledge of the language model while letting the EEG supply evidence before the model decides.

Meeting this challenge requires a view of what the EEG could supply. A sentence is not a flat string of tokens: words carry lexical meaning, belong to grammatical categories and combine into phrases and sentences at characteristic timescales. Although these processes run largely automatically, brain signals reflect them. Cortical activity tracks words, phrases and sentences at distinct rhythms (Ding et al., 2016). Imaging, electrophysiological and single-neuron studies relate activity to word order, part of speech and phrase construction (Pallier et al., 2011; Nelson et al., 2017; Cai et al., 2026). Event-related potentials distinguish closed-class (function) from open-class (content) words (Brown et al., 1999) and index lexical-semantic processing in the N400 (Kutas & Federmeier, 2011), and non-invasive word decoding is shaped by part of speech and word length (d'Ascoli et al., 2025). Most of this evidence, however, comes from averaged responses, controlled presentation or invasive recordings. Does the EEG of a single fixation in natural reading carry usable lexical and grammatical evidence, or is such information visible only in averages? Either answer matters, because it shows which decisions a decoder can base on the EEG and which it should leave to sequence structure and the language model.

Fixation-related potentials (FRPs) make this question testable in natural reading. Time-locking the EEG to the onset of each fixation yields word-level brain responses without controlled stimulus presentation (Dimigen et al., 2011), so the recordings used for decoding can be examined for word-class differences, for single-fixation discriminability and for overlap between the responses to successive words. If lexical and grammatical evidence is weak and graded, a decoder may do better not to commit to a word from one fixation but to keep lexical and grammatical evidence separate, combine them across the sentence and leave contextual adjudication to the language model. Structured prediction offers such a combination: a conditional random field (CRF) scores whole sequences from local evidence and learned transitions between neighbouring labels (Lafferty et al., 2001). A semantic head, a part-of-speech (POS) head and a joint word–POS CRF may therefore turn weak EEG evidence into a restricted candidate set that a language model scores in context, at lower cost than LLM candidate generation, and this study tests this design.

The two heads also need a contextual representation of each fixated word, and the FRP analysis bears on what kind. If the responses to successive words overlap, nearby words matter more than distant ones. Transformers model all pairwise interactions but lack a locality bias and need large training sets (Dosovitskiy et al., 2021), whereas Hyena operators and their StripedHyena 2 arrangement mix context through short, medium and long convolutions with limited attention (Poli et al., 2023; Ku et al., 2025). An encoder with an explicit locality bias may therefore suit fixation sequences, and we test whether the trained encoder uses nearby fixations in the way the overlap of their responses suggests.

To test these ideas, we characterise word-level EEG in natural reading with fixation-related analyses and build EEGHyena2, a decoder in which a bidirectional encoder inspired by StripedHyena 2 feeds a semantic head and a POS head, a joint word–POS CRF scores a restricted lattice, and a frozen Llama-2 scores the candidates in context. We test it on two natural-reading EEG corpora, ZuCo (Hollenstein et al., 2018, 2020) and ROAMM (Sun et al., 2026), against the published state of the art, EEG foundation models and a length-matched language-model-only control. Section 2 reviews the literature, Section 3 reports the neural analyses, Section 4 describes the decoder and the experimental setup, Section 5 presents the results and Section 6 concludes.

# 2. Literature Review

## 2.1. Lexical and grammatical structure: from brain signals to brain-to-text decoding

Sentence comprehension involves word meaning, grammatical category and hierarchical structure, and each leaves traces in brain activity. When syllables, phrases and sentences are presented at 4, 2 and 1 Hz, cortical responses track all three rates (Ding et al., 2016), and activity in language areas increases with the size of syntactic constituents (Pallier et al., 2011; Nelson et al., 2017). In story reading, event-related potentials show a late frontal negativity for closed-class but not open-class words (Brown et al., 1999), and the N400 indexes lexical-semantic processing (Kutas & Federmeier, 2011). Single-neuron recordings relate activity to the building blocks of language (Cai et al., 2026), and in natural reading the EEG around a fixation carries information about orthographic, phonological and contextual features (Raghavan & Parra, 2025). Semantic relatedness and prediction effects can also be separated by EEG classification (Trammel et al., 2023).

Brain-to-text decoding has drawn on this structure mostly through language models. Neural language decoding spans semantic-gist reconstruction from fMRI (Tang et al., 2023), retrieval of perceived speech segments from MEG and EEG (Défossez et al., 2023), identification of the listened language from EEG (Ariza et al., 2025), visual-content decoding from EEG aligned with multimodal models (Jing et al., 2025), word decoding (d'Ascoli et al., 2025) and sentence reconstruction. EEG-to-text decoding of natural reading has moved from sequence-to-sequence translation with BART (Wang & Ji, 2022) to contrastive and masked EEG–text pretraining (Wang et al., 2024), subject-dependent encoders with language-based refinement (Amrani et al., 2026) and LLM-guided candidate selection (Zheng et al., 2026). Along this line the language model has taken on a growing share of the decision, but evaluation has had to keep pace: teacher forcing at test time inflates scores, noise-input controls reveal weak dependence on neural input in some systems (Jo et al., 2024), and benchmark design affects teacher-forcing-free inference (Zhang et al., 2026). A supplied first word or the output length can also raise overlap without EEG, because BLEU includes a brevity penalty (Papineni et al., 2002) and output length can be controlled directly (Kikuchi et al., 2016).

A few studies have used lexical or grammatical structure more directly. Part of speech can be decoded from EEG recorded during reading, with Transformers outperforming linear classifiers on single words and word pairs (Murphy et al., 2022), and the predictions of large-scale non-invasive word decoders carry part-of-speech and word-length information even when the decoded word is wrong (d'Ascoli et al., 2025). In sentence decoding, structure enters mainly through the language model: invasive speech neuroprostheses decode phonemes and assemble sentences with a language model (Willett et al., 2023), fMRI decoders let a language model propose continuations that an encoding model scores (Tang et al., 2023), and recent EEG-to-text methods guide an LLM with recovered semantic keywords (Yang et al., 2026) or with sentence-level targets such as topic, sentiment, length and surprisal (Wang et al., 2026). These lines have remained separate. Grammatical properties have been decoded as stand-alone classification targets, sentence decoders have left grammar to the language model, and we found no EEG-to-text decoder that combines lexical and grammatical evidence from the EEG in the sequence decision itself. Moreover, most of the neural evidence comes from averaged responses or controlled presentation, whereas a decoder must act on one fixation of one reader.

### 2.2. Fixation-related potentials and encoders for fixation sequences

A decoder that acts on single fixations depends on what the EEG of one fixation carries. Natural reading has no stimulus onsets: the eyes sample words at a self-paced rate, some words are skipped, and the features used by current EEG-to-text models are aggregated over each word's fixations (Hollenstein et al., 2018; Wang & Ji, 2022; Zheng et al., 2026). Three properties of these features matter for the design: whether lexical and grammatical classes differ in them, whether the differences hold for single fixations, and how far the response to one word extends into those of its neighbours.

FRPs were developed for this setting. Early FRP studies time-locked the EEG to fixation onsets to examine parafoveal processing in reading (Baccino & Manunta, 2005), and co-registration of eye tracking and EEG made it possible to recover classic word-level ERP effects during natural sentence reading (Dimigen et al., 2011). FRP studies have since shown an N400 predictability effect time-locked to the first fixation on a word, without an accompanying frequency effect (Kretzschmar et al., 2015), and visual and orthographic effects of parafoveal previews (Degno et al., 2019). Because fixations follow one another within a few hundred milliseconds, overlapping responses and ocular artefacts have required dedicated correction methods (Ehinger & Dimigen, 2019). Skipping, which depends on word length, frequency and predictability (Rayner, 1998), removes some words from the record altogether, and the decision to skip can precede full identification of the word (Milligan & Schotter, 2026), while preview lets a word influence processing before it is fixated (Schotter et al., 2012).

These studies show that fixation-locked EEG reflects word properties during natural reading, but they leave two questions open for decoding. FRP research tests group-level effects of controlled manipulations of target words, so it does not show whether a word property can be read from a single fixation; and EEG-to-text studies use fixation-aligned features without characterising what they carry. Analysing the decoding corpus with FRPs and single-fixation classification links the two, showing which word properties the decoder can expect in its input and how strongly the responses to neighbouring words overlap.

The answer also bears on the encoder that supplies the contextual representation of each fixated word. Transformers compute content-dependent interactions between all positions through self-attention (Vaswani et al., 2017), but they lack the locality bias of convolutions and generalise poorly when training data are limited (Dosovitskiy et al., 2021). EEG models have addressed this by combining convolution with attention, as in EEGNet (Lawhern et al., 2018), EEG Conformer (Song et al., 2023), and multi-band spatio-temporal encoders for visual decoding (Xue et al., 2026), or by pretraining foundation models such as LaBraM and NeuroLM on thousands of hours of EEG (Jiang et al., 2024, 2025). Hyena operators offer a third option: they replace attention with long convolutions whose filters are generated implicitly and gated by the input (Poli et al., 2023), and StripedHyena 2 interleaves short, medium and long convolution operators with attention (Ku et al., 2025). This arrangement matches the structure of fixation sequences. Its short- and medium-range filters favour neighbouring words, where overlapping responses and preview effects arise, while long filters and an attention layer still reach distant context. With at most 56 fixated words per ZuCo recording, the reason to adopt it is this inductive bias, but whether an encoder trained on fixation-linked EEG uses neighbouring words in this way has not been examined, so we build the decoder on a StripedHyena-2-inspired encoder and inspect what it learns.

# 3. Neural Evidence: What Fixation-Linked EEG Carries About Words

This section asks what the EEG of one fixation carries about the word being read. We examine whether readers sample content and function words differently, whether fixation-linked EEG and the band-power input of the decoder differ between the two classes on average, whether a single fixated word carries class information, and how far the responses to successive words overlap. The analyses use ZuCo, whose release provides continuous sentence EEG, first-fixation onsets and eye-tracking measures for every word, and they are independent of the decoder.

## 3.1. ZuCo recordings, word classes and word-level features

We use the multi-subject setting of Zheng et al. (2026), which combines the normal-reading tasks of ZuCo 1.0 (Hollenstein et al., 2018) (sentiment reading, SR, of movie reviews; normal reading, NR, of Wikipedia sentences) with the normal-reading task of ZuCo 2.0 (Hollenstein et al., 2020). EEG was recorded with a 128-channel EGI HydroCel net at 500 Hz together with an EyeLink eye tracker. The released word-level features give, for each fixated word, the band power, a frequency-domain summary of the scalp EEG, in eight frequency bands ($\theta_1$ 4–6 Hz, $\theta_2$ 6.5–8 Hz, $\alpha_1$ 8.5–10 Hz, $\alpha_2$ 10.5–13 Hz, $\beta_1$ 13.5–18 Hz, $\beta_2$ 18.5–30 Hz, $\gamma_1$ 30.5–40 Hz, $\gamma_2$ 40–49.5 Hz) at each of 105 electrodes, computed over the word's gaze duration; we denote this vector $f_t \in \mathbb{R}^{840}$.

Gold POS tags were obtained by tagging the reference sentences with spaCy (Honnibal et al., 2020) using the 17 Universal POS tags and aligning reference tokens to EEG word positions by character spans; all EEG words were aligned. Eye-tracking measures (number of fixations, first-fixation duration, gaze duration and total reading time) were read from the same release for the analyses in Section 3.2; the released durations are given in samples at 500 Hz, which we verified against the length of each word's first-fixation EEG segment and converted to milliseconds. For the fixation-related analyses we cut epochs from $-200$ to $+800$ ms around the onset of each word's first fixation from the continuous sentence EEG (104 recorded channels), subtracted the mean of the pre-fixation interval and downsampled the epochs to 100 Hz.

We distinguish two broad word classes. Content words carry most of the lexical meaning of a sentence: nouns, proper nouns, verbs, adjectives and adverbs (Universal POS tags NOUN, PROPN, VERB, ADJ and ADV). Function words mainly signal grammatical relations: determiners, adpositions, pronouns, auxiliaries, conjunctions, particles and all remaining tags. In the ZuCo sentence of Fig. 1, "achieves", "ultimate" and "comedy" are content words and "in", "it's" and "the" are function words. In this recording the reader looked longer at content than at function words (mean gaze duration 498 vs 173 ms) and skipped three tokens, "II", "--" and "as", for which no word-level EEG exists. Word class is the coarsest grammatical distinction the decoder could target, and it is the property the following analyses test.

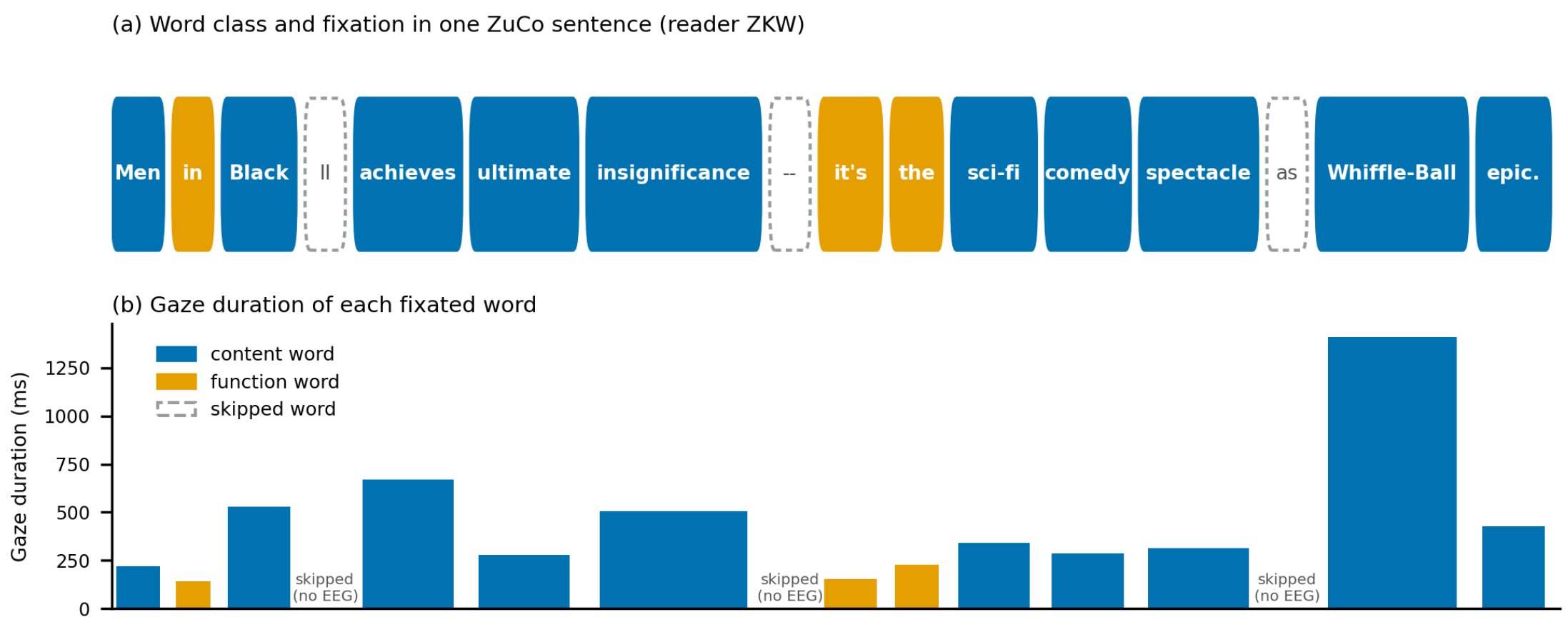

**Fig. 1.** Content and function words in one ZuCo test recording (reader ZKW, sentiment-reading task; the recording analysed in Fig. 12). (a) Fixated content words (blue), fixated function words (orange) and skipped words (dashed outline), which have no word-level EEG. (b) Gaze duration of each fixated word. Content words are nouns, proper nouns, verbs, adjectives and adverbs; function words carry all other Universal POS tags.

## 3.2. Fixation-related analyses

**Reading behaviour and FRPs.** To characterise what word-level signals carry, we analyse all 165,907 EEG-aligned words of the three splits with MNE-Python (Gramfort et al., 2013), computing statistics across the 30 readers from reader-level means. Words are labelled as content or function words (Section 3.1). We compare skipping rates and gaze durations between the two classes, compute FRPs (epochs with a peak-to-peak amplitude below 150 $\mu$V; 160,190 epochs) and test the content–function difference from 0 to 800 ms with a spatio-temporal cluster permutation test (Maris and Oostenveld, 2007) (cluster-forming threshold $p < 0.01$, 2,000 permutations).

**Spectral band-power contrasts.** For the 840-D model input we compute, per band and electrode, a $t$-test across readers of the content–function difference with false-discovery-rate correction (Benjamini and Hochberg, 1995).

**Single-word class information.** To measure how much class information a single word carries, we train within-reader logistic-regression classifiers (5-fold cross-validation grouped by sentence) on fixation-locked amplitudes, band-power features or the four eye-tracking measures, and average $k \in \{1,2,\dots,64\}$ randomly drawn words of the same class into each example; the area under the ROC curve (AUC) as a function of $k$ shows how quickly the signal emerges from noise. We also report linear probes that predict word properties from each input, and a probe of whether the EEG recorded at word $t + k$ ($k = -3,\dots,3$) carries information about word $t$ beyond the neighbour's own class, length and frequency.

**Overlap between successive fixations**. For each word we measured the delay from its first-fixation onset to the first-fixation onsets of the next three words, which shows how far the epoch of one word extends into the responses to its neighbours.

## 3.3. Findings

Readers treated content and function words differently (Fig. 2). They skipped 46.4% of function words and 13.3% of content words, and the difference held at equal word length. Content words were fixated longer (median gaze duration 240 vs 205 ms, averaged over readers; $t(29) = 11.3$, $p < 10^{-11}$), and gaze duration decreased with word frequency. Skipped words, which have no EEG, are mostly function words, which is the case the skipped-word insertion of the decoder addresses (Section 4.7).

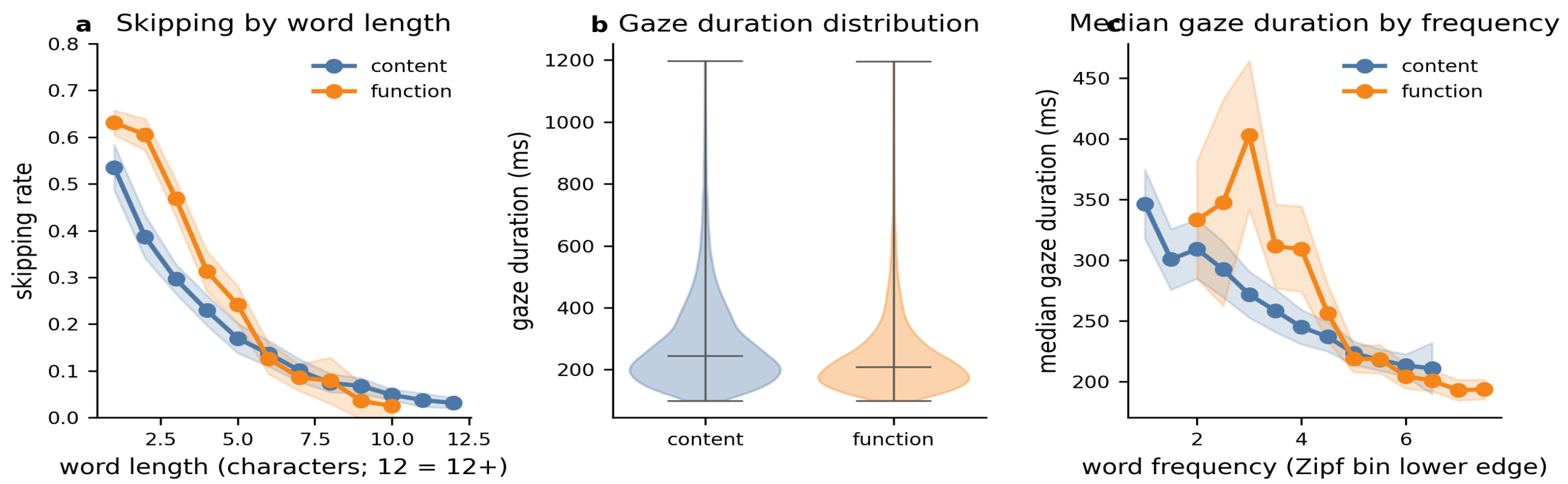

**Fig. 2.** Eye movements by word class (30 readers). (a) Skipping rate by word length (95% CI). (b) Distribution of gaze duration. (c) Median gaze duration by word frequency (Zipf scale; mean and 95% CI across readers).

The first fixation on the next word followed that on the current word by a median of 306 ms, and 71% of these onsets fell inside the 0–800 ms epoch of the current word; for the word after that, the median delay was 626 ms (59% within the epoch). Responses to neighbouring words therefore overlap within each word's epoch.

The EEG also differed between the classes on average (Fig. 3). After the visual lambda response that peaks about 80 ms after fixation onset, content words evoked a more negative response over posterior electrodes, and a cluster permutation test identified one significant spatio-temporal cluster from 90 to 290 ms over 71 of 104 channels ($p = 0.0005$). In the 840-D model input, every band contained electrodes with a significant content–function difference after false-discovery-rate correction (7 to 35 of 104 electrodes per band; Fig. 4): content words had relatively more posterior $\theta_1$ power and less central and frontal $\alpha$, $\beta$ and $\gamma$ power. Some significant electrodes lay at the frontal and temporal edges, where ocular and muscle activity contribute. A regression on word surprisal (Smith and Kutas, 2015), controlling for frequency and length, showed an early effect (0–230 ms) but no significant effect in the N400 window (300–500 ms).

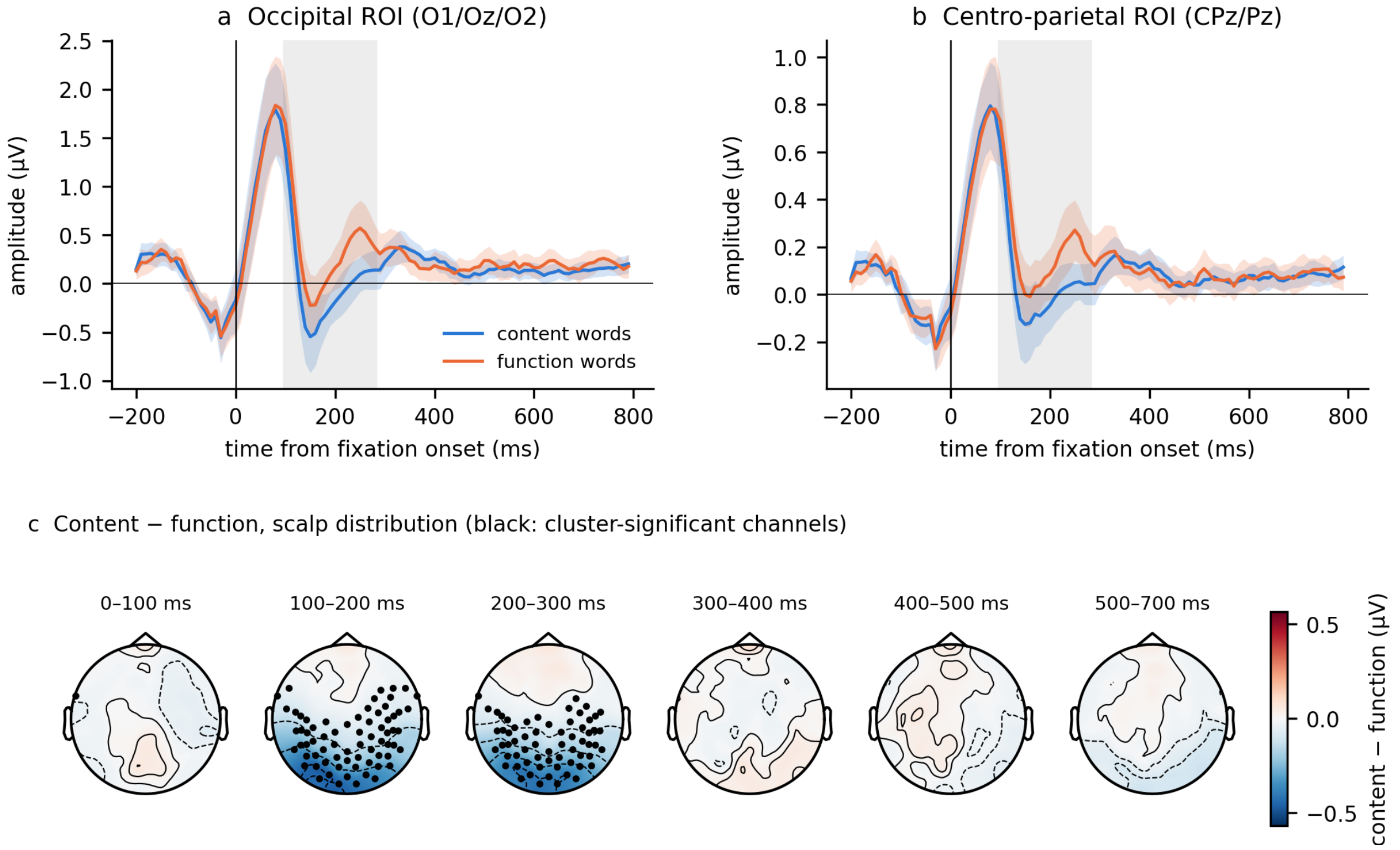

**Fig. 3.** FRPs for content and function words (160,190 epochs, 30 readers). (a, b) Grand averages at occipital and centro-parietal regions (mean and 95% CI across readers); grey shading marks the significant cluster. (c) Scalp distribution of the content–function difference; black dots mark channels in the significant cluster.

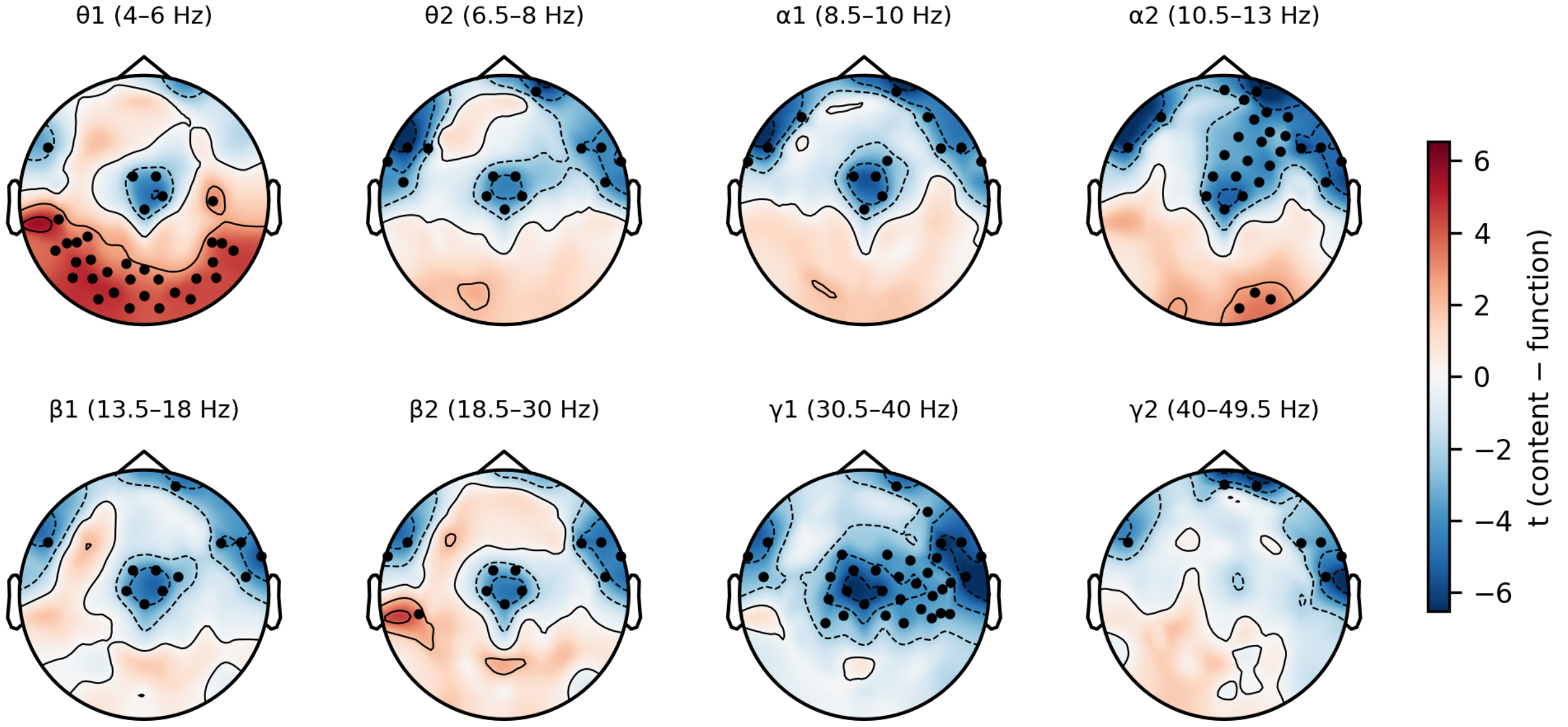

**Fig. 4.** Content–function difference in the 840-D model input: $t$ values across 30 readers per band and electrode; black dots mark electrodes significant after false-discovery-rate correction.

Within each reader, fixation-locked EEG classified content and function words with a median AUC of 0.53 for single words, and the band-power features reached the same 0.53, whereas the four eye-tracking measures reached 0.64. Averaging 64 words of the same class raised the AUC to 0.78 (fixation-locked EEG) and 0.80 (band power), against 1.00 for eye tracking. Linear probes on the 840-D features showed the same pattern: they identified the reader with a balanced accuracy of 0.905 (30 readers) but predicted word class with 0.518 and word length, frequency and surprisal with correlations of 0.10–0.13, below the eye-tracking measures (0.59; 0.34, 0.33 and 0.28). Added to eye tracking, the EEG features improved prediction reliably only for surprisal ($\Delta r = 0.016$, 95% CI [0.005, 0.026]) and position in the sentence. The EEG recorded at neighbouring words carried no detectable information about word $t$ beyond the neighbours' own properties, whereas the eye-tracking measures at the next word did (AUC gain 0.019, $p = 2 \times 10^{-6}$), the spillover effect known from reading research (Rayner, 1998).

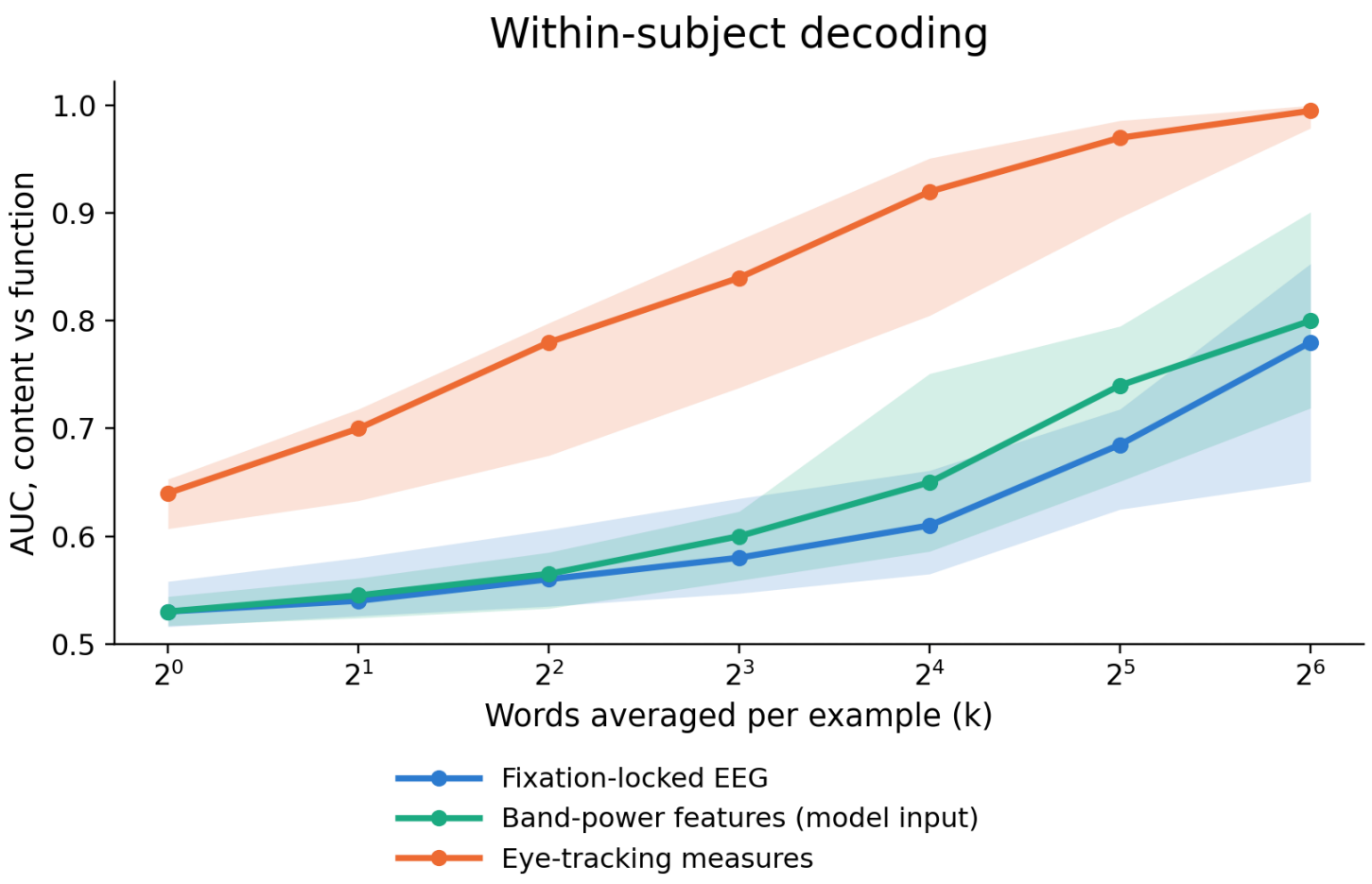

**Fig. 5.** Within-reader classification of content versus function words from single words ($k = 1$) and from averages of $k$ words of the same class (median and interquartile range over 29 readers).

### 3.4. Implications for the decoder design

These findings give a qualified answer to the question raised in the Introduction. Word class leaves a reliable trace in fixation-linked EEG, so the grammatical distinction most relevant to sentence construction is present in the decoder's input; at the level of one fixation, however, the trace is too weak to support a decision, and the band-power features say more about who is reading than about what is read, a form of inter-subject variability that also limits cross-subject generalisation in other EEG decoding tasks (Apicella et al., 2024). A decoder should therefore not commit to a word or a class from one fixation. It should carry graded evidence forward, with a semantic head that ranks a lexicon and a POS head that outputs a distribution over categories (Section 4.4), and let sequence structure decide by combining this evidence with word-specific tag priors and learned transitions in a joint word–POS CRF (Section 4.6).

The remaining findings shape the rest of the search. The clearest EEG signal beyond eye tracking related to surprisal, but contextual predictability is the quantity a language model estimates directly, so a frozen language model scores the candidates in context within the same search instead of generating them (Section 4.7). Skipped words leave no EEG and are mostly function words, so a small insertion vocabulary and a training-derived length policy can fill these positions without neural evidence. Finally, the responses to successive words overlapped in time, although the band-power features carried no detectable information about neighbouring words. An encoder that favours nearby fixations is therefore plausible but not guaranteed to help, and Section 5.3 examines which distances the trained encoder uses.

# 4. Methodology

Our decoder applies structured prediction to EEG-to-text decoding. Before large language models, language technology reduced uncertainty in stages, with probabilistic lexical access, part-of-speech tagging and syntactic disambiguation narrowing the hypotheses before a final decision (Jurafsky, 1996; Ratnaparkhi, 1996; Pota et al., 2019). Linear-chain CRFs score a complete label sequence from local emissions and learned transitions with global normalisation (Lafferty et al., 2001), neural encoders have been combined with CRF output layers and attention for sequence labelling (Lample et al., 2016; Sun et al., 2020; Lin et al., 2021), constrained decoding builds lexical requirements into search (Lu et al., 2021), and shallow fusion adds language-model probabilities to a task model's scores during beam search (Gulcehre et al., 2017), and beam-search variants that promote semantic diversity improve language generation (Na et al., 2025). We use these methods for a division of labour opposite to that of LLM4EEG2Text: EEG-conditioned lexical scores, aligned with a pretrained text space (van den Oord et al., 2018; Devlin et al., 2019), and POS evidence define a restricted lattice of (word, POS) states, a CRF scores paths through it, and the language model scores only the surviving candidates (Fig. 6). Because the search runs over a bounded lattice instead of repeated open-vocabulary generation, it also reduces large-model computation, whose cost depends on generated tokens, batching and decoding strategy (Fernandez et al., 2025; Oviedo et al., 2026). Section 4.1 defines the task, Sections 4.2–4.7 describe the decoder and Section 4.8 the experimental setup.

## 4.1. Problem formulation and evaluation protocol

A participant reads a sentence $y = (y_1, \ldots, y_N)$ of $N$ words while EEG and eye movements are recorded. The reader fixates $T \leq N$ of the words; the remaining words are skipped. Each fixated word $t = 1, \ldots, T$ is represented by a word-level EEG feature vector, giving the input $x_{1:T}$. The decoder must produce a text $\hat{y}$ that is close to $y$:

$$\hat{y} = \mathrm{arg}max_{y' \in y(p)} \, \mathrm{score}(y' \mid x_{1:T}, p), \tag{1}$$

where $p$ is the text prefix given by the evaluation protocol. We follow the known-first-word protocol of Zheng et al. (2026): $p$ is the reference text up to the end of the first fixated word, and the evaluated output contains this prefix. The protocol therefore measures the reconstruction of the remaining words given the first one. All methods in this study, including the language-model-only control, receive the same prefix.

## 4.2. Model overview

Our model (Fig. 6) maps the sequence $x_{1:T}$ to two kinds of word-level evidence, a semantic embedding and a POS distribution, combines them in a restricted lattice of candidate (word, POS) states, scores state sequences with a joint word–POS CRF (Lafferty et al., 2001), and decodes the sentence with a beam search in which a frozen Llama-2-7B (Touvron et al., 2023) scores every word in context.

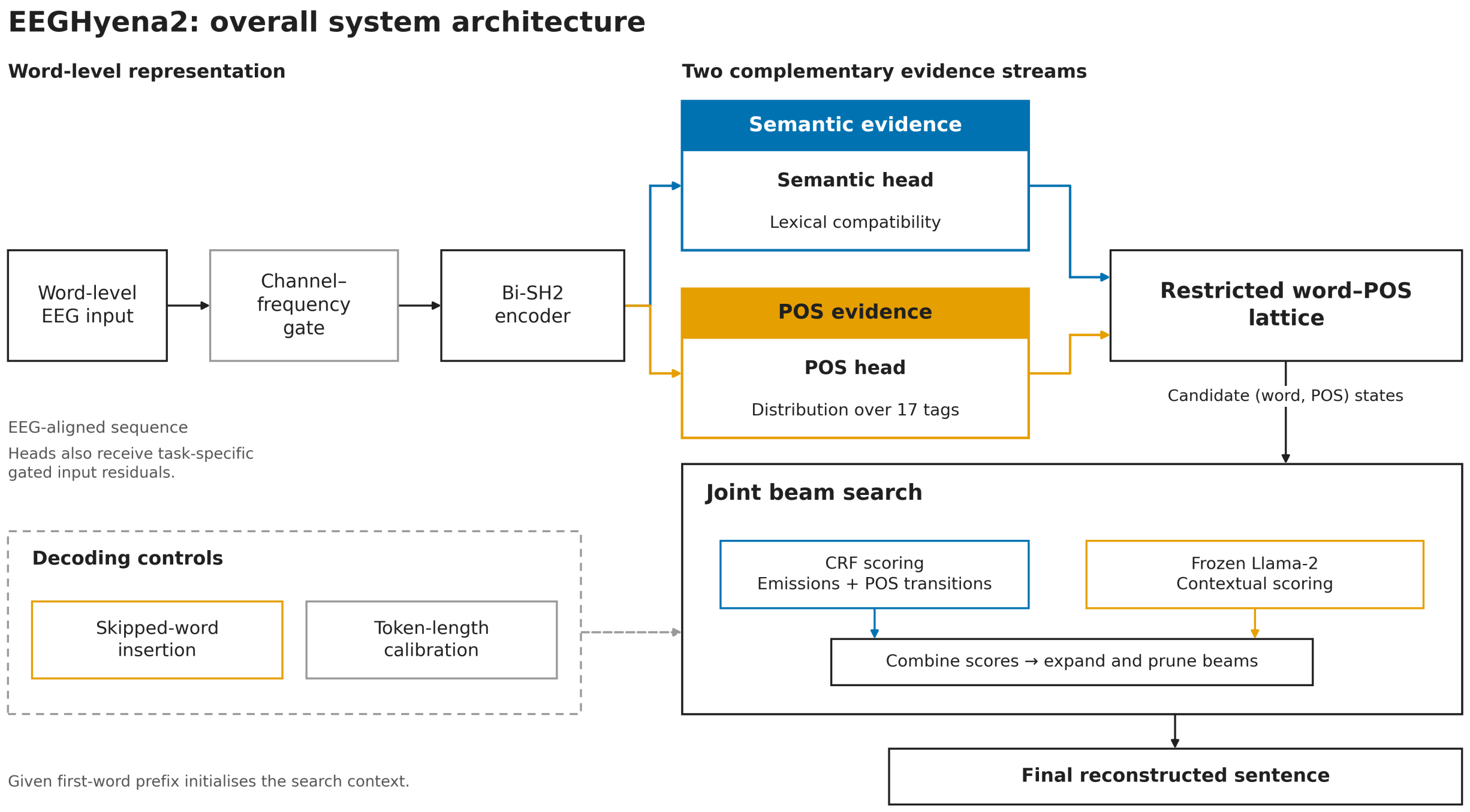

**Fig. 6.** Overall system architecture. The word-level EEG encoder supplies semantic and POS evidence to a restricted word–POS lattice. CRF transitions and frozen Llama-2 contextual scores contribute within joint beam search; skipped-word recovery and token-length calibration constrain the same search.

The semantic head, the POS head and the joint word–POS CRF (Sections 4.4–4.6) form the evidence interface that this study proposes. The channel–frequency gate and the Bi-SH2 encoder (Section 4.3) supply word-level representations to both heads, and the frozen Llama-2, skipped-word insertion and length calibration (Section 4.7) complete the search. Because the roles are separated, each part can be replaced or controlled: Section 5 substitutes LaBraM features for the band-power input, training-set averages for the EEG-derived emissions and a length-matched language-model-only continuation for the whole EEG-conditioned decoder.

## 4.3. EEG feature encoding

### 4.3.1. Channel–frequency gating

As in the reference implementation, each band-power vector is z-normalised over its own entries,

$$x_t = \frac{f_t - \mu(f_t)}{\sigma(f_t)}, \tag{2}$$

so that the model receives the relative band–electrode profile of each word.

The 840 features have a band-by-electrode structure. A learnable gate assigns a weight in $(0,1)$ to every band–electrode pair,

$$G = \sigma(\Lambda) \in (0,1)^{8\times 105}, \qquad \tilde{x}_t = x_t \odot \mathrm{vec}(G), \tag{3}$$

where $\sigma$ is the logistic function and $\Lambda$ is initialised to 4, so that the gate starts close to pass-through ($G \approx 0.982$). The gate re-weights inputs multiplicatively, in the spirit of squeeze-and-excitation channel weighting (Hu et al., 2018), but is static per feature, so that a trained gate can be read as a statement about each band and electrode. A shared gate feeds the encoder, and two task-specific gates $G^{\mathrm{sem}}$ and $G^{\mathrm{pos}}$ feed residual connections into the two heads (Section 4.4). The pass-through initialisation retains all features initially, and their subsequent weights are learned from the task objectives.

#### 4.3.2. Bidirectional StripedHyena-2-inspired encoder (Bi-SH2)

Bi-SH2 is a model of contextual mixing across word positions. A causal convolution uses preceding positions only; applying the same stack to the valid sequence and to its reversal gives information from both reading directions. The final attention layer makes interactions between positions content dependent. The three operator types arranged in six residual blocks, shared weights, sequence masking and maximum length of 56 are held as specified below for reproducibility.

The design follows Section 3.3. Successive fixations overlap within each word's epoch, and preview lets neighbouring words influence processing (Schotter et al., 2012), so short and medium filters favour nearby words, whereas long filters and a final attention layer allow distant context. Compared with a Transformer encoder of the same width, the convolutional operators add an explicit locality bias, which suits a training set of about 134,000 EEG-aligned words (Section 4.8.1).

The gated inputs are projected and combined with a learned position embedding, $h_t^{(0)} = W_{\mathrm{in}}\tilde{x}_t + p_t$ with $W_{\mathrm{in}} \in \mathbb{R}^{384\times 840}$. The encoder (Fig. 7) stacks gated-convolution operators of the Hyena family (Poli et al., 2023) in the short–medium–long pattern of StripedHyena 2 (Ku et al., 2025). An operator maps $u \in \mathbb{R}^{T\times d}$ to

$$[q, k, v] = \phi\big(uW_{qkv}\big), \qquad y = \big(q \odot (h * (k \odot v))\big)W_o, \tag{4}$$

where $\phi$ is a causal depthwise convolution of length 3 and $h\,*$ a causal depthwise long convolution, $(h * z)_t = \sum_{\ell\geq 0} h_\ell \odot z_{t-\ell}$, whose filter differs by operator type. The short explicit (SE) operator learns five free taps; the medium regularised (MR) operator learns 15 taps multiplied by a learnable exponential decay; the long implicit (LI) operator generates up to 64 taps with a small network over sinusoidal position features, modulated by a fixed exponential decay; all long convolutions are applied by FFT.

Each operator sits in a pre-normalised residual block with a channel-mixing MLP, and six blocks are used in the order SE, MR, LI, SE, MR, LI.

Because the operators are causal, bidirectional context is obtained by applying the same stack $\mathcal{F}$ (shared weights) to the sequence and to its reversal within the valid prefix, $\mathcal{R}$, as in bidirectional recurrent networks (Schuster and Paliwal, 1997):

$$h^{\rightarrow} = \mathcal{F}(h^{(0)}), \quad h^{\leftarrow} = \mathcal{R}\Big(\mathcal{F}\big(\mathcal{R}(h^{(0)})\big)\Big), \quad h = \mathrm{LN}\big([h^{\rightarrow}; h^{\leftarrow}]W_f\big). \tag{5}$$

Inputs are masked inside every operator to restrict computation to valid word positions. One non-causal self-attention layer (Vaswani et al., 2017) with eight heads follows, and a final projection gives the contextual word representation $z_t \in \mathbb{R}^{768}$. Our implementation uses lightweight Hyena-inspired operators tailored to sequences of up to 56 words.

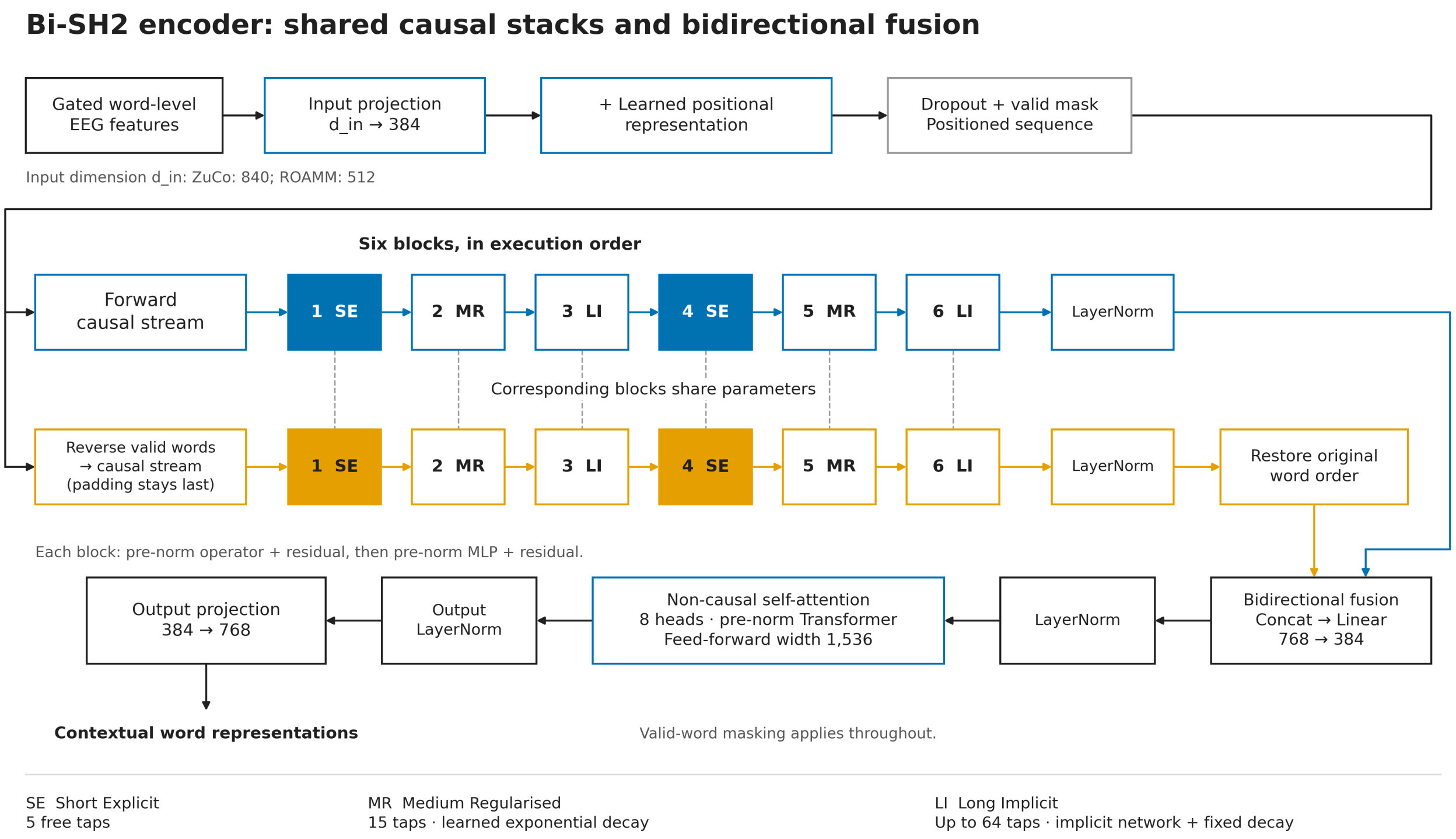

**Fig. 7.** Bi-SH2 word-context encoder. Forward and valid-word-reversed streams use the same six SE–MR–LI–SE–MR–LI causal blocks with shared weights. The reversed output is returned to the original word order before bidirectional fusion, non-causal self-attention and output projection.

Our encoder code projects the gated 840-D input to 384 dimensions, adds a learned position embedding (64-position allocation), applies a six-block shared-weight SE–MR–LI–SE–MR–LI stack to the valid forward and reversed sequences, reverses backward outputs within valid lengths, and fuses the two streams by a 768-to-384 linear layer followed by LayerNorm. A pre-norm, eight-head Transformer encoder layer (feed-forward width 1,536) follows; the output LayerNorm and projection map to the 768-D semantic/POS representation. The main configuration uses dropout 0.1 and obtains bidirectional overall context from the two causal-filter streams.

## 4.4. Semantic and POS heads

Part of speech (POS) denotes a broad grammatical class such as noun, verb or determiner. The 17 Universal POS labels express grammatical uncertainty at a coarse level. In the present design, a semantic embedding judges the lexical compatibility of a candidate word, whereas the POS output helps distinguish syntactically different possibilities.

The two heads follow from Section 3.4. Because the EEG of a single word carries weak and graded information, neither head is asked to name a word or a class. The semantic head ranks every lexicon entry by compatibility with the EEG representation, so the decoder keeps a candidate set rather than a single guess. The POS head outputs a full distribution over the 17 tags, so weak class evidence can be outweighed by tag priors, transitions and linguistic context during search. Keeping the two outputs separate also makes each inspectable and replaceable, which the emission-replacement control in Section 4.8.4 uses.

Each head combines the contextual representation with a gated residual path from its own input features and applies a task-specific linear projection and a two-layer MLP.

The semantic head outputs a 768-dimensional unit-normalised embedding, and the POS head outputs log-probabilities over the 17 Universal tags.

## 4.5. Inference-safe lexicon and candidate lattice

The restricted lattice is a list of possible (word, POS) states at each EEG-aligned position.

The candidate vocabulary $\mathcal{V}$ contains the training-set words and the whole-word entries of the BERT vocabulary (Devlin et al., 2019) ($|\mathcal{V}| = 23{,}890$). Each word $w$ is represented by $e_w$, the BERT-base-uncased hidden state of its last word piece when the word is encoded in isolation. The reference implementation derives its training targets from BERT states conditioned on the reference prefix, encoding words in isolation makes the training target and the inference candidate the same vector. The semantic score of word $w$ at position $t$ is a calibrated log-probability

$$a_t(w) = \log \frac{\exp(s_t^{\top} e_w/\tau)}{\sum_{v \in \mathcal{V}} \exp(s_t^{\top} e_v/\tau)}, \tag{6}$$

with a learned temperature $\tau$ (initialised to 0.1). For each position the $K = 20$ highest-scoring words are retained; each is paired with at most three POS tags, the most frequent ones observed for it in the training data (unseen-in-training words use the POS head's highest-probability tags). A state $q = (w, z)$ receives the emission score

$$\psi_t(w, z) = a_t(w) + \pi_t(z) + c(w, z), \qquad c(w, z) = \log \frac{n(w,z)+0.1}{\sum_{z'} n(w,z')+1.7}, \tag{7}$$

where $c(w, z)$ is the smoothed training-set probability of tag $z$ for word $w$ ($n$: tag counts over the 17 tags). The language-model score is also added during decoding.

The inference lexicon uses training words and permitted whole-word BERT entries encoded in isolation. Gold (word, POS) states are inserted into training lattices for the CRF likelihood; inference candidates are ranked from the fixed lexicon.

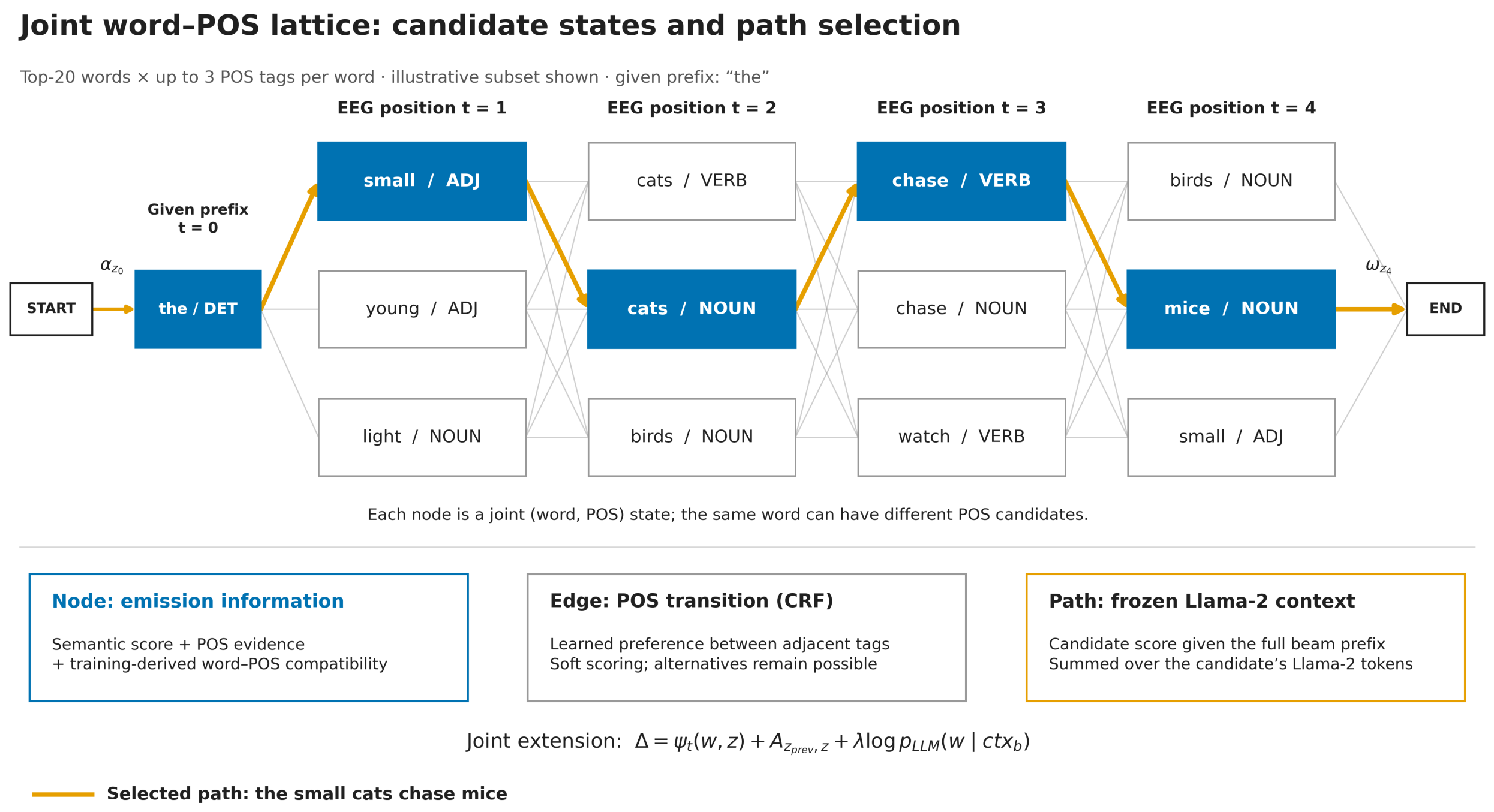

**Fig. 8.** Illustrative restricted word–POS lattice. Each node represents a candidate (word, POS) state at an EEG-aligned position; node emissions combine lexical and POS evidence, CRF transitions provide learned soft preferences, and frozen Llama-2 supplies prefix-dependent context scores within beam search. The highlighted path is illustrative.

## 4.6. Joint word–POS CRF and training objective

A conditional random field (CRF) is a globally normalised sequence-scoring model. It scores entire paths through the candidate lattice using local EEG-conditioned emissions and learned adjacent POS transitions. During inference, the CRF contribution is combined incrementally with Llama-2 scores in the same beam search.

A linear-chain CRF scores a state sequence $q_{1:T} = (w_t, z_t)_{t=1}^{T}$ over the restricted lattice,

$$S(q_{1:T}) = \alpha_{z_1} + \textstyle\sum_{t=1}^{T} \psi_t(w_t, z_t) + \displaystyle\sum_{t=2}^{T} A_{z_{t-1} z_t} + \omega_{z_T}, \tag{8}$$

with a learned $17 \times 17$ POS transition matrix $A$ and start and end vectors $\alpha, \omega$. The joint objective is the negative log-likelihood of the gold path, normalised by the number of words,

$$L_{\text{joint}} = \frac{1}{T}\left(\log \sum_{q'_{1:T}} e^{S(q'_{1:T})} - S(q^{*}_{1:T})\right), \tag{9}$$

where the partition function is computed by the forward algorithm. Because the likelihood is undefined when the gold state is absent from the lattice, the gold state is inserted into training lattices only. A globally normalised CRF scores whole sequences and avoids the label bias of locally normalised models (Lafferty et al., 2001). Transitions are modelled between POS tags rather than between words because a $17 \times 17$ matrix can be estimated reliably from the training data, whereas word-to-word transitions would be sparse. Dependencies between words are left to the language model during decoding.

The semantic objective $L_{\text{sem}}$ is the dual contrastive alignment loss of the reference implementation.

The model combines the baseline-compatible EEG–BERT alignment loss and the word–POS CRF negative log-likelihood. Alignment includes positive/negative cosine terms and a clean/noisy-view InfoNCE term (contrastive temperature 0.3); the noisy view replaces individual EEG features with standard-normal noise at probability 0.3. Learned log-scale uncertainty weights, initialized at zero, balance the objectives.

$$L_{\text{sem}} = 4\left(1 - \frac{1}{N}\sum_i \cos\left(s_i, e_i^+\right)\right) + \frac{2}{N^2}\sum_{i,j} \cos\left(s_i, e_j^-\right) + \text{CE}\left(\left[\cos(s_i, s_j')/T_c\right]_{ij}, i\right), \tag{10}$$

over the $N$ valid words of a batch, where $e_i^+$ is the embedding of the gold word, $e_j^-$ the embedding of a corrupted word (word pieces randomly replaced), $s_j'$ the semantic output for an input in which 30% of the feature entries are replaced by standard normal noise, and $T_c = 0.3$; the last term is an InfoNCE objective (van den Oord et al., 2018) that aligns clean and noisy views of the same word. Matching the contrastive objective to the reference holds this part of training constant. The two objectives are combined with learned log-scale weights,

$$L_{\text{total}} = e^{-\sigma_{\text{sem}}} L_{\text{sem}} + \sigma_{\text{sem}} + e^{-\sigma_{\text{joint}}} L_{\text{joint}} + \sigma_{\text{joint}}, \tag{11}$$

initialised at $\sigma = 0$. This form follows the uncertainty weighting of Kendall et al. (2018); it balances the two loss scales through learned weights. All components except the frozen language models are trained jointly.

## 4.7. Language-model fusion, skipped-word recovery and length calibration

**Fused scoring.** Decoding proceeds over EEG positions $t = 1, \dots, T-1$ (position 0 is fixed to the given first word). A beam $b$ holds a partial word sequence and the Llama-2 context $\text{ctx}_b$ (the given prefix followed by the words decoded so far). Extending $b$ with lattice state $(w, z)$ at position $t$ adds

$$\Delta = \psi_t(w, z) + A_{z_{\text{prev}}z} + \lambda \log p_{\text{LLM}}(w \mid \text{ctx}_b), \tag{12}$$

$\log p_{\text{LLM}} \lambda = 1$where sums the log-probabilities of the word's Llama-2 tokens. This is shallow fusion of a language model into a structured decoder (Gulcehre et al., 2017); we fix as an untuned design choice. Only the 30 states with the highest emission are expanded per position, and 10 beams are kept.

**Skipped-word insertion.** ZuCo provides EEG only for fixated words, whereas the reference contains every word; in the training set readers skip 0.35 words per fixated word, mostly short function words. Between two EEG positions the beam may therefore insert up to two words from a gap vocabulary $\mathcal{G}$, the 50 words most frequently skipped in the training set (64.5% of skipped tokens on the development set), plus the punctuation marks "," and ".". Each insertion is scored by $\lambda \log p_{\text{LLM}}$ and receives a bonus $\beta$; the decoder excludes immediate repetition of the same word. A beam's total score is

$$S_b = S_{\text{CRF}}(b) + \lambda\, S_{\text{LLM}}(b) + \beta\, n_{\text{ins}}(b), \tag{13}$$

where $n_{\text{ins}}$ is the number of inserted words.

**Token-level length calibration.** Corpus BLEU penalises outputs that are shorter than the references and loses precision when they are longer (Papineni et al., 2002), and the reference evaluation computes BLEU on Llama-2 tokens. We therefore budget the output length in tokens. With a target of $L^* = \text{round}(T \cdot \rho_{\text{ref}})$ tokens, an insertion of word $w$ at position $t$ is allowed only if

$$|\text{ctx}_b| + |\text{tok}(w)| + (T - t)\, \rho_{\text{eeg}} \leq L^*, \tag{14}$$

where $\rho_{\text{ref}} = 2.245$ is the number of reference Llama-2 tokens per EEG word and $\rho_{\text{eeg}} = 1.599$ the number of Llama-2 tokens per decoded EEG word, both measured on the training set; the term $(T - t)\rho_{\text{eeg}}$ reserves room for the EEG words still to be decoded. The budget is computed from the number of fixated words $T$ and training-set length statistics. The output is the highest-scoring beam. We refer to this decoder as EEGHyena2; its predecessors are evaluated in Section 5.2.2: CRF 10-best reranking combines CRF hypotheses with Llama-2 and dependency-parse checks; LM-fused decoding searches directly with Llama-2 probabilities (with or without skipped-word insertion); and the word-budget variant limits insertions by a training-derived word budget $\text{round}(0.53\,T)$, the training rate of skipped words and punctuation per EEG word.

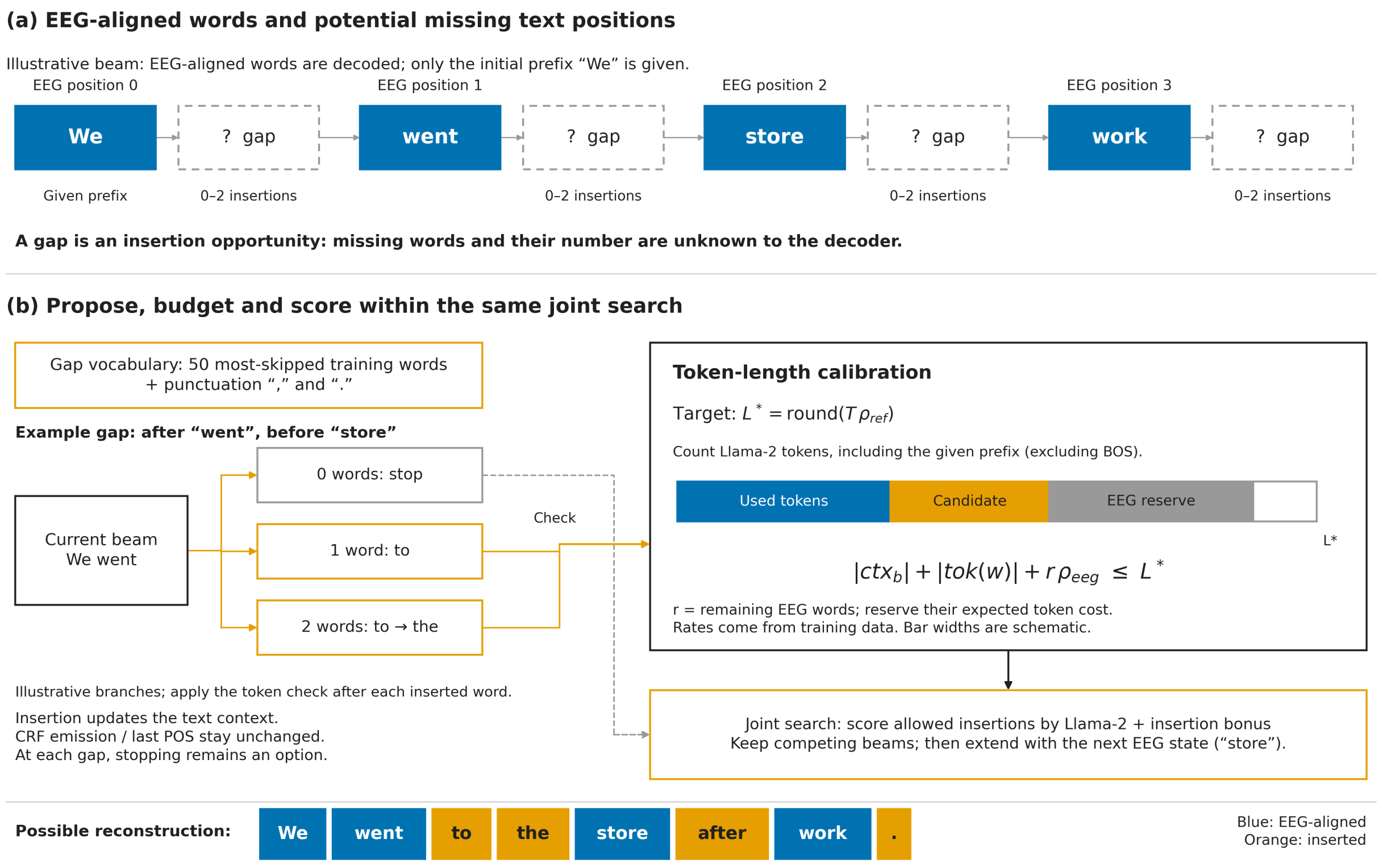

**Fig. 9.** Skipped-word recovery and token-length calibration in joint beam search. (a) The decoder receives EEG-aligned positions but does not know the missing words or their number; each gap permits up to two optional insertions. (b) Gap-word candidates from the training vocabulary are scored in context and admitted only if the Llama-2 token-length budget preserves capacity for remaining EEG words.

## 4.8. Experimental setup

Both corpora were used with the same protocol. This section describes the data splits, training, comparison systems, metrics and the analyses of the trained decoder.

### 4.8.1. Datasets and splits

The decoder was trained and tested separately on two English natural-reading corpora, ZuCo and ROAMM, with the same task definition, decoder design and control logic. The corpora differ in recording system, number of electrodes, text material and splitting unit; the corpus-specific inputs and splits are given below and in Table 1.

For ZuCo, the split follows the reference implementation: within each task and for every reader, the first 80% of sentence indices form the training set, the next 10% the development set and the remaining 10% the test set.

Sentence recordings were removed when sentence-level EEG was missing or invalid, when word features were missing or had the wrong dimension, or when fewer than the minimum number of words carried EEG, following the reference filters. Table 1 summarises the resulting data: 9,704 / 1,173 / 1,195 sentence recordings with 134,111 / 15,105 / 16,691 EEG-aligned words from 30 readers and 984 distinct sentences, with at most 56 EEG words per recording. Forty-eight sentences occur in more than one task and five appear in more than one split, as in the reference split.

**Table 1.** Retained sentence recordings and EEG-aligned word occurrences by corpus and split. SR: sentiment reading; NR: normal reading. ROAMM was split by article page.

| Task | Readers | Sentences per reader | Training | Development | Test | Total |
|---|---|---|---|---|---|---|
| ZuCo 1.0 SR (movie reviews) | 12 | 400 | 3,386 | 418 | 354 | 4,158 |
| ZuCo 1.0 NR (Wikipedia) | 12 | 300 | 2,323 | 315 | 315 | 2,953 |
| ZuCo 2.0 NR (Wikipedia) | 18 | 349 | 3,995 | 440 | 526 | 4,961 |
| ZuCo total | 30 | – | 9,704 | 1,173 | 1,195 | 12,072 |
| ROAMM total | 44 | varies | 11,565 | 1,557 | 1,206 | 14,328 |
| ZuCo EEG-aligned words | – | – | 134,111 | 15,105 | 16,691 | 165,907 |
| ROAMM EEG-aligned words | – | – | 160,451 | 21,777 | 16,142 | 198,370 |

ROAMM (Sun et al., 2026) is a natural-reading corpus with its own recording layout, text material and repeated readings of article pages. It provides 64 EEG channels, so the eight-band channel–frequency input contains 512 features. The EEGHyena2, LaBraM and NeuroLM-B pipelines were adapted to this input and trained or fine-tuned on ROAMM; the rest of the decoder was unchanged. The source ROAMM corpus includes 44 participants; EEG was acquired with 64-channel BioSemi ActiveTwo at 2,048 Hz and downsampled to 256 Hz during the published preprocessing pipeline (Sun et al., 2026).

To limit text leakage across repeated readings, article pages rather than individual EEG records were assigned to an approximately 80/10/10 split. The resulting material contained 11,565 training, 1,557 development and 1,206 test records. The emission-replacement analysis (Section 4.8.4) used a paired subset of 320 test records scored under the same known-first-word condition; these records represented 47 distinct stimulus sentences and 27 distinct first words.

The 512-dimensional ROAMM word features were derived from the preprocessed continuous EEG sampled at 256 Hz. For each valid first-pass fixation (at least 0.1 s and shorter than 3 s), continuous-run EEG from each of the 64 channels was filtered into eight frequency bands (4–6, 6.5–8, 8.5–10, 10.5–13, 13.5–18, 18.5–30, 30.5–40 and 40–49.5 Hz) using fourth-order Butterworth bandpass filters. For each band and channel, the Hilbert amplitude envelope was averaged over the samples within that fixation. When a word received multiple qualifying fixations, the eight-by-64 features were averaged across fixations with weights proportional to fixation duration. Each resulting 512-dimensional word vector was z-normalised across its entries.

### 4.8.2. Training and implementation details

Table 2 lists the settings. The model has 13.0M trainable parameters. AdamW was used with a learning rate of $5 \times 10^{-4}$, weight decay 0.01, batch size 256, 30 epochs and seed 888, matching the second-stage settings of the reference implementation. For the reported EEGHyena2 decoding we use the final-epoch last.pt checkpoint. The insertion bonus was chosen from $\{2, 4, 8\}$ by BLEU-1 on 200 development recordings ($\beta = 4$); no parameter was tuned on the test set.

**Table 2.** Model and training settings of the proposed decoder (EEGHyena2).

| Setting | Value |
|---|---|
| Input | 840 band-power features per fixated word, z-normalised per word |
| Encoder | width 384; blocks SE–MR–LI ×2; filter lengths 5 / 15 / ≤64; shared forward/reversed weights |
| Attention | 1 layer, 8 heads; output width 768 |
| Heads | gated residual semantic head (768-D, unit norm); POS head (17 tags) |
| Lexicon | 23,890 words; BERT-base-uncased isolated-word embeddings; temperature $\tau$ learned (init 0.1) |
| Lattice | top $K = 20$ words × at most 3 POS tags per position |
| Decoding | Llama-2-7B (bfloat16), $\lambda = 1$, beam 10, 30 states per position, $\beta = 4$ |
| Gap vocabulary | 50 most-skipped training words + "," and "."; at most 2 insertions per gap |
| Length budget | $\rho_{\text{ref}} = 2.245$, $\rho_{\text{eeg}} = 1.599$ Llama-2 tokens per EEG word (training set) |

#### 4.8.3. Comparisons

**LLM4EEG2Text (Zheng et al., 2026).** LLM4EEG2Text is the strongest published method for multi-subject ZuCo decoding under the known-first-word protocol and represents LLM candidate generation with EEG reranking. We adopt its data selection, split, input features, Llama-2 tokenisation and metrics, so our ZuCo results are directly comparable with its reported values. In its report it exceeded CET-MAE (Wang et al., 2024), a contrastive and masked EEG–text pretraining model, and a variant that chooses among the LLM candidates at random, while an oracle given the ground-truth semantics set an upper bound; we list these published values in Table 3. We also re-ran the authors' code (commit 6f6949a) in the same setting with two patches required by our system (data-loader workers set to zero because of a shared-memory limit, and a guard for a missing field in the ZuCo 2.0 files). We reproduced the method on a subset of 669 test recordings, the first in its processing order and all from the ZuCo 1.0 tasks, and compare it with our decoder on exactly these recordings. On ROAMM, the reproduced model (the original Transformer in Table 5) serves as the reference method.

**LaBraM (Jiang et al., 2024).** LaBraM represents EEG foundation models that learn transferable representations by masked prediction of neural tokens, and it tests whether large-scale EEG pretraining improves the decoder. LaBraM-Base (5.8M parameters, pretrained on about 2,500 h of EEG) is used as a frozen feature extractor in place of the 840-D input, retaining the downstream EEGHyena2 encoder and decoder. For each fixated word, a 1-s window starting at the onset of its first fixation is cut from the continuous sentence EEG (98.4% of training words; the remainder use the concatenated fixation segments), restricted to 68 ZuCo electrodes selected using the approximate HydroCel-to-10–10 correspondence documented by Luu and Ferree (2005), resampled to 200 Hz and scaled as in LaBraM. The mean-pooled 200-D output replaces $x_t$. The insertion bonus was chosen on the development set as for EEGHyena2 ($\beta = 8$).

**NeuroLM-B (Jiang et al., 2025).** NeuroLM couples an EEG tokenizer with a GPT-2 language model (254M parameters, pretrained on about 25,000 h of EEG). It represents EEG–language foundation models that generate text end to end and tests whether one generative model can replace the structured decoder. We fine-tuned the released NeuroLM-B checkpoint with the authors' instruction-tuning recipe on the instruction "Question: What sentence was the subject reading? Answer:" with the reference sentence as answer. The input is the sentence EEG (68 channels, 200 Hz, 1-s patches), truncated at 13 s by the model's 1,024-token context. Training uses AdamW with a peak learning rate of $5 \times 10^{-4}$ and cosine decay, an effective batch of 64 and five epochs, optimizing the EEG-to-text instruction task. At test time the given prefix is appended after "Answer:" and decoding is greedy. We report the checkpoint with the lowest development loss, reached after the first epoch.

**Length-matched Llama-2 control (no EEG).** Because the protocol supplies the first word and the decoder uses the fixation count to set the output length, an EEG-free language model given the same information can already overlap with the reference (Jo et al., 2024). To measure this overlap, the same frozen Llama-2-7B continues the given prefix. A token budget from Eq. 14 limits the continuation to at most $L^{*} - |\mathrm{tok}(p)|$ tokens, so it receives the same length information as EEGHyena2. Generation uses five-beam search with 3-gram repetition blocking, and outputs are lowercased to match the references.

#### 4.8.4. Evaluation metrics and statistical inference

A generated sentence may share many common words with a reference yet change their order, omit content words or paraphrase its meaning, so we report several metrics. BLEU is sensitive to matched token sequences and output length; ROUGE-1 reports lexical overlap, and ROUGE-L adds word order. BERTScore compares contextual word representations, character and word error rates (CER, WER) measure the edit burden, and role-F1 compares automatically parsed predicate–argument relations. Whole-string and post-prefix scores are kept distinct because the first word is supplied by the protocol.

Following the reference implementation, we compute corpus BLEU-1 to BLEU-4 (Papineni et al., 2002) on Llama-2 tokens, ROUGE-1 recall, precision and F1 (Lin, 2004), BERTScore F1 (Zhang et al., 2020) against the BERT-decoded reference. Corpus BLEU-$n$ combines modified $n$-gram precisions $p_k$ with a brevity penalty.

where $r$ and $c$ are the total reference and output lengths. We add four measures. ROUGE-L F1 (Lin, 2004) credits the longest common subsequence and therefore word order. The character error rate,

$$\mathrm{CER} = \frac{\sum_i \mathrm{edit}(\hat{y}_i, y_i)}{\sum_i |y_i|}, \tag{15}$$

sums character-level edit distances over the test set and divides by the number of reference characters. Role-F1 is the micro-averaged F1 of predicate–argument relations extracted with the spaCy dependency parser (Honnibal et al., 2020), computed after removing relations that lie entirely inside the given prefix,

$$\text{Role-}F_1 = \frac{2PR}{P+R}, \quad P = \frac{\sum_i |\mathcal{R}(\hat{y}_i) \cap \mathcal{R}(y_i)|}{\sum_i |\mathcal{R}(\hat{y}_i)|}, \quad R = \frac{\sum_i |\mathcal{R}(\hat{y}_i) \cap \mathcal{R}(y_i)|}{\sum_i |\mathcal{R}(y_i)|}, \tag{16}$$

where $\mathcal{R}(\cdot)$ is the multiset of (predicate, role, argument) triples. Content-word ROUGE-1 applies ROUGE-1 after removing function words (spaCy English stop words) and punctuation from output and reference, over the whole evaluated string and over the part after the given prefix. Methods are compared on identical test recordings. Uncertainty is estimated with a paired bootstrap over test sentences (Efron and Tibshirani, 1993): the same 2,000 resamples are applied to both methods, and we report 95% percentile intervals of each metric and of the difference between methods.

In addition, two analyses examine the trained decoder. We inspect the learned CRF transitions, gates and encoder, and, on ROAMM, we replace the EEG-derived emissions of the frozen decoder with training-set averages.

For the encoder we report the mean absolute kernel weight of each operator type by lag, the self-attention weight by signed word distance relative to uniform attention, and the effective receptive field (Luo et al., 2016): for output word $t$ and input word $j$,

$$\mathrm{ERF}(t,j) = \frac{\left\| \nabla_{x_j} \| P_{z_t} \|_2^2 \right\|_1}{\sum_{j'} \left\| \nabla_{x_{j'}} \| P_{z_t} \|_2^2 \right\|_1} \tag{17}$$

where $P$ holds four fixed random projections of the output, aggregated over 291 eligible test sentences from 300 selected for analysis and grouped by $j - t$.

To separate the fitted structured decoder from the evidence in the current EEG record, we froze the ROAMM EEGHyena2 decoder (checkpoint, lexicon, POS/CRF transitions, first-word prefix, number of EEG-aligned positions, Llama-2 model, gap insertion, beam search and token-length policy). We replaced the semantic-head vector at each EEG position by its mean over training-set word positions (11,565 training records) and the POS log-probabilities by the log of the training-mean POS distribution. This removes the content of the test EEG from both emissions but keeps the supplied prefix and the EEG-derived position count.

The emission replacement uses these 320 test records (47 distinct stimuli) and scores only the continuation after the supplied first word, with ROUGE-L F1 computed by a regex-word evaluator. Intervals use a stimulus-cluster bootstrap because the corpus contains repeated readings of the same stimuli.

# 5. Results

## 5.1. Benchmark performance and inference cost

### 5.1.1. ZuCo: reconstruction quality and decoding cost

Our reproduction of LLM4EEG2Text reached BLEU-1/2/3 of 21.7/13.0/9.1 on the first 120 test recordings, close to the published 21.3/11.9/7.7 on the full test set. We reproduced the method on a subset of 669 recordings, all from ZuCo 1.0, so we use the published values for the full test set and the reproduction for a paired comparison on this subset (see Table 3).

On the full test set, EEGHyena2 matched the published LLM4EEG2Text on BLEU-1 to BLEU-3 (21.4/12.1/7.8 vs 21.3/11.9/7.7) and exceeded it by 3.0 points in ROUGE-1 F1 (23.3 vs 20.3), with both higher recall and higher precision; its BERTScore F1 was lower (81.5 vs 84.8; Table 3). On the 669 recordings decoded by the reproduction, EEGHyena2 was higher at every BLEU order and in ROUGE-1 F1 and role-F1, and lower in BERTScore F1. EEGHyena2 decoded a test recording in 2.47 s, against 8.5–19 s for the reproduced LLM4EEG2Text that scores 300 LLM candidates per step.

**Table 3.** Decoding results under the known-first-word protocol. The first block reproduces Table 3 of Zheng et al. (2026) (values as published); the Oracle LLM selects candidates with the ground-truth semantics and is an upper bound, the Random LLM selects candidates at random. "All" is the full test set (1,195 recordings). The last block pairs the reproduction with EEGHyena2 on the same 669 recordings. Role-F1 excludes relations inside the given prefix. Words: mean output length (reference length in parentheses).

| **Method** | **Records** | **BLEU-1/2/3/4** | **ROUGE-1 R/P/F1** | **BERTScore F1** | **Role-F1** | **Words (ref.)** |
|---|---|---|---|---|---|---|
| CET-MAE (Wang et al., 2024) (reported) | paper | 18.1 / 9.7 / 6.2 / – | 17.8 / 18.1 / 17.9 | 83.5 | – | – |
| Random LLM (reported) | paper | 14.6 / 9.0 / 6.3 / – | 16.1 / 18.3 / 16.5 | 83.5 | – | – |
| LLM4EEG2Text (reported) | paper | 21.3 / 11.9 / 7.7 / – | 19.8 / 20.8 / 20.3 | 84.8 | – | – |
| Oracle LLM, upper bound (reported) | paper | 42.9 / 33.1 / 26.8 / – | 46.1 / 47.6 / 47.0 | 88.9 | – | – |
| **Ours (EEGHyena2)** | all | **21.4** / 12.1 / 7.8 / 5.5 | 21.1 / **27.9** / **23.3** | 81.5 | **2.3** | 27.4 (21.3) |
| Ours + LaBraM (frozen) | all | 21.6 / 12.0 / 7.8 / 5.4 | 20.9 / 26.4 / 22.5 | 81.4 | 2.1 | 27.6 (21.3) |
| NeuroLM-B | all | 20.4 / 11.7 / 7.9 / 5.6 | 19.3 / 23.6 / 19.6 | 83.8 | 1.7 | 18.7 (21.3) |
| LLM only, length-matched (no EEG) | all | 20.8 / 12.3 / 8.5 / 6.0 | 19.1 / 24.7 / 21.0 | 82.7 | 1.9 | 18.2 (21.3) |
| LLM4EEG2Text (reproduction) | 669 | 19.8 / 11.0 / 7.2 / 5.0 | 17.7 / 17.8 / 17.5 | 82.9 | 1.1 | 21.4 (20.7) |
| **Ours (EEGHyena2)** | 669 | **20.6** / **11.8** / **7.7** / **5.2** | 20.8 / 27.0 / **22.6** | 81.8 | **2.9** | 27.1 (20.7) |

Frozen LaBraM features in place of the band-power input changed the scores by less than one point (BLEU-1 21.6 vs 21.4; ROUGE-1 F1 22.5 vs 23.3). NeuroLM-B produced shorter outputs with lower BLEU-1 and ROUGE-1 F1 and a higher BERTScore F1. Against NeuroLM-B, EEGHyena2 had a higher role-F1 and ROUGE-L F1 and an 11.7-point lower CER, all with bootstrap intervals excluding zero (see Table 4).

**Table 4.** Relation, order and character-level measures on the full test set with 95% bootstrap intervals (2,000 paired resamples over test sentences). The last column gives the difference of each method from EEGHyena2 (method − EEGHyena2); for CER lower is better.

| Method | Role-F1 | ROUGE-L F1 | CER |
|---|---|---|---|
| **Ours (EEGHyena2)** | **2.33** [1.90, 2.76] | **21.81** [21.27, 22.37] | 73.84 [73.17, 74.58] |
| Ours + LaBraM: difference | -0.18 [-0.61, +0.27] | -0.82 [-1.24, -0.43] | -0.07 [-0.42, +0.30] |
| NeuroLM-B: difference | -0.66 [-1.13, -0.17] | -3.11 [-3.61, -2.62] | +11.69 [10.10, 13.31] |
| LLM only, length-matched: difference | -0.44 [-1.02, +0.13] | -1.58 [-2.09, -1.03] | -0.91 [-1.46, -0.39] |

### 5.1.2. ROAMM: reconstruction quality and decoding cost

On 669 matched ROAMM test records (Table 5), EEGHyena2 exceeded the reproduced LLM4EEG2Text Transformer on BLEU-2 to BLEU-4, ROUGE-1 F1 and ROUGE-L F1 and had a lower CER (75.3% vs 84.7%). NeuroLM-B was 0.61 ROUGE-L points higher than EEGHyena2 (paired bootstrap $p = 0.048$), but had a substantially higher CER and shorter outputs. LaBraM features again changed the scores little.

**Table 5.** Cross-model ROAMM results on 669 matched test records. BLEU, ROUGE and CER are percentages; WER is a proportion.

| Model | BLEU-1 ↑ | BLEU-2 ↑ | BLEU-3 ↑ | BLEU-4 ↑ | ROUGE-1 F1 ↑ | ROUGE-2 F1 ↑ | ROUGE-L F1 ↑ | CER ↓ |
|---|---|---|---|---|---|---|---|---|
| **EEGHyena2** | 19.88 | **8.44** | 3.57 | 1.80 | **21.82** | **4.07** | 19.59 | 75.30 |
| LLM4EEG2Text | 20.29 | 7.19 | 3.17 | 1.70 | 20.87 | - | 18.99 | 84.69 |
| LaBraM + EEGHyena2 | 19.40 | 8.26 | 3.48 | 1.78 | 21.24 | 3.96 | 19.26 | 74.38 |
| NeuroLM-B | 19.08 | 8.05 | 3.90 | 2.20 | 21.67 | 3.77 | 20.21 | 88.68 |

Decoding took 3.34 s per ROAMM record for EEGHyena2 and 33.70 s for the reproduced LLM4EEG2Text.

## 5.2. Decoder ablations and length-controlled baselines

### 5.2.1. ZuCo: length-matched language-model-only control

Given the same prefix and token budget but no EEG, Llama-2 scored lower than EEGHyena2 in ROUGE-L F1 (difference 1.58 points, 95% CI [1.03, 2.09]) and in whole-string content-word ROUGE-1 F1 (1.50 points, [1.15, 1.86]), where EEGHyena2's precision was higher (Tables 4 and 6). The control was higher in BLEU-2 to BLEU-4 and BERTScore F1, had a lower CER and recovered more content words after the given prefix (0.48 points, [0.10, 0.87]). Beyond the prefix, every method recovered fewer than 3% of the reference content words.

**Table 6.** Content-word overlap (ROUGE-1 after removing function words and punctuation), full test set. "Whole" scores the evaluated string including the given prefix; "after prefix" scores only the decoded continuation. Differences are method − EEGHyena2 with 95% paired bootstrap intervals.

| Method | Whole: P / R / F1 | Whole: F1 difference | After prefix: F1 | After prefix: F1 difference |
|---|---|---|---|---|
| **Ours (EEGHyena2)** | **21.6** / 9.2 / **12.3** | – | 1.4 | – |
| Ours + LaBraM | 20.2 / 9.4 / 12.1 | -0.21 [-0.48, -0.06] | 1.4 | +0.02 [-0.26, +0.32] |
| NeuroLM-B | 13.2 / 10.7 / 10.9 | -1.43 [-1.99, -0.85] | 2.6 | +1.14 [0.58, 1.72] |
| LLM only, length-matched | 13.1 / 9.6 / 10.8 | -1.50 [-1.86, -1.15] | 1.9 | +0.48 [0.10, 0.87] |

The two systems reach similar overlap by different routes. Given only the prefix and a length budget, Llama-2 writes fluent, generic continuations that share short n-grams and contextual meaning with the reference often enough to raise BLEU-2 to BLEU-4 and BERTScore; these scores are therefore partly available without any neural input. EEGHyena2 restricts each EEG position to lexicon entries ranked by the semantic head and orders them under learned POS transitions with inserted function words, and its advantage lies in the measures that reward words in reference order (ROUGE-L) and in precision. This precision gain does not come from recovering more content: after the prefix, the language model and NeuroLM-B recovered slightly more reference content words than EEGHyena2, and all three stayed near floor. Because EEGHyena2 nonetheless had the higher overall ROUGE-1 precision (27.9% vs 24.7%), its additional matches were mostly function words and the given prefix, and it produced fewer unsupported content words.

### 5.2.2. ZuCo: structured-search and length-budget ablation

Skipped-word insertion and length budgets produced the largest gains (Table 7, Fig. 10a). CRF best-path decoding and 10-best reranking reached a BLEU-1 of about 12, and Llama-2 fusion restricted to EEG positions gave 10.1 with short outputs. Adding skipped-word insertion raised BLEU-1 to 17.1, a word budget to 19.1 and a token budget to 21.4.

**Table 7.** Decoder variants on the full test set (1,195 recordings). Word accuracy is the share of EEG positions whose decoded word equals the gold word; it includes the given first word. "Known gaps" uses the true skipped-word count from eye tracking as an additional oracle input.

| Variant | Decoding | BLEU-1/2/3/4 | ROUGE-1 R/P/F1 | BERT Score F1 | Word acc. | Words |
|---|---|---|---|---|---|---|
| CRF best-path decoding | CRF best path, no LLM | 12.1 / 7.8 / 5.5 / 4.0 | – | 81.4 | 7.79 | – |
| CRF + LM 10-best reranking | CRF 10-best + LLM rerank | 12.4 / 8.0 / 5.6 / 4.1 | 16.8 / 25.1 / 19.9 | 81.4 | 7.81 | – |
| LM fusion (EEG words only) | fused, EEG words only | 10.1 / 6.6 / 4.7 / 3.5 | 17.8 / 45.0 / 24.3 | 79.9 | 9.54 | 14.8 |
| LM fusion + skipped-word insertion | fused + skipped words | 17.1 / 10.2 / 6.8 / 4.8 | 20.3 / 35.8 / 24.3 | 80.4 | 9.30 | 32.1 |
| LM fusion + known gap positions | fused, known gaps | 13.7 / 8.7 / 6.1 / 4.4 | 19.4 / 41.3 / 24.8 | 79.7 | 9.47 | 20.4 |
| LM fusion + word budget | + word budget | 19.1 / 11.1 / 7.4 / 5.2 | 20.7 / 28.6 / 23.5 | 82.2 | 8.61 | 20.9 |
| EEGHyena2 (token budget) | + token budget | **21.4** / **12.1** / **7.8** / **5.5** | 21.1 / 27.9 / 23.3 | 81.5 | 8.69 | 27.4 |

The ablation shows that, on this benchmark, BLEU follows output length more closely than word identity. Fusing Llama-2 into the search over EEG positions gave the highest word accuracy (9.54%) and ROUGE-1 precision (45.0%) but the lowest BLEU-1, because outputs of 14.8 words against 21.3 in the references incurred the brevity penalty. Each later step lengthened or calibrated the output and raised BLEU-1 by 2–7 points, while word accuracy at the EEG positions fell slightly, from 9.54% to 8.69%. The search policy therefore decides how many words are produced and which function words fill the gaps, whereas the identity of the words at EEG positions is set mainly by the emissions and stayed between 7.8% and 9.5% in all variants. Supplying the true number of skipped words from eye tracking lowered BLEU-1 relative to free insertion (13.7 vs 17.1), so the insertion gain reflects output length and plausible function words rather than recovery of the skipped words themselves. Reranking ten CRF hypotheses with Llama-2 added only 0.3 BLEU-1 points: a language model applied after the search cannot repair candidate sets and lengths that the search has already fixed, and it contributes most when it scores candidates inside the search.

### 5.2.3. ROAMM: emission replacement

Replacing the semantic and POS emissions of the frozen decoder with their training-set averages reduced ROUGE-L F1 from 10.47% to 7.89% (difference +2.58 [+1.22, +3.97] points) at nearly the same output length (23.90 vs 23.70 words; Fig. 10b).

With the lattice, CRF transitions, gap insertion, length policy and language model held fixed and the output length almost unchanged, removing the test-record content from the semantic and POS emissions lowered ROUGE-L F1 by about a quarter. The structured search alone therefore does not account for the output of the decoder: the evidence computed from each test record changes which words it selects and how it orders them. Two limits apply. Both emission channels were replaced together, so the analysis does not separate the semantic from the POS contribution, and the test records cover only 47 distinct stimuli, which is reflected in the width of the stimulus-cluster interval. Read together with the ZuCo content-word results (Section 5.2.1), this evidence is likely to act more on the form of the output than on the recovery of content words.

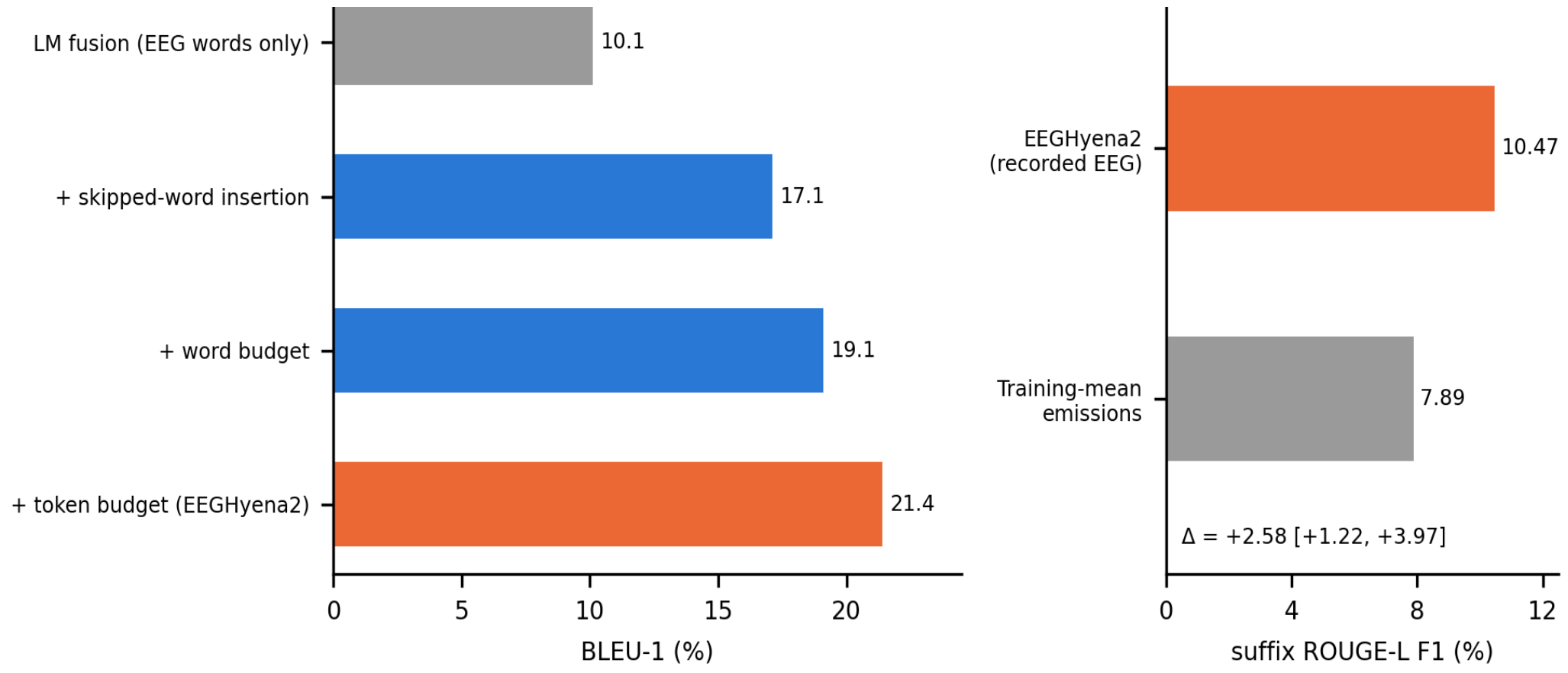

**Fig. 10.** Distinct decoding and evidence-control comparisons. (a) Within-pipeline ZuCo progression from fused EEG-word decoding to skipped-word insertion and word/token length budgets (BLEU-1; Table 7). (b) ROAMM suffix-only ROUGE-L F1 for 320 UIDs: EEGHyena2 with recorded EEG and with training-mean semantic and POS emissions in the frozen decoder; Δ is their difference with its 95% stimulus-cluster bootstrap interval.

### 5.3. What the structured components learned

#### 5.3.1. Learned word–POS transitions

In terms of CRF transitions, the joint CRF learned POS transitions that reflect English syntax (Fig. 11). The highest transition scores were particle→verb (“to go”), auxiliary→verb, pronoun→auxiliary, adjective→noun, determiner→adjective and adposition→determiner (0.41–0.45); the lowest were adverb→noun, auxiliary→noun, particle→noun and adjective→verb (−0.44 to −0.42). The end scores favoured sentences ending in nouns, proper nouns and numbers and penalised endings in conjunctions and particles.

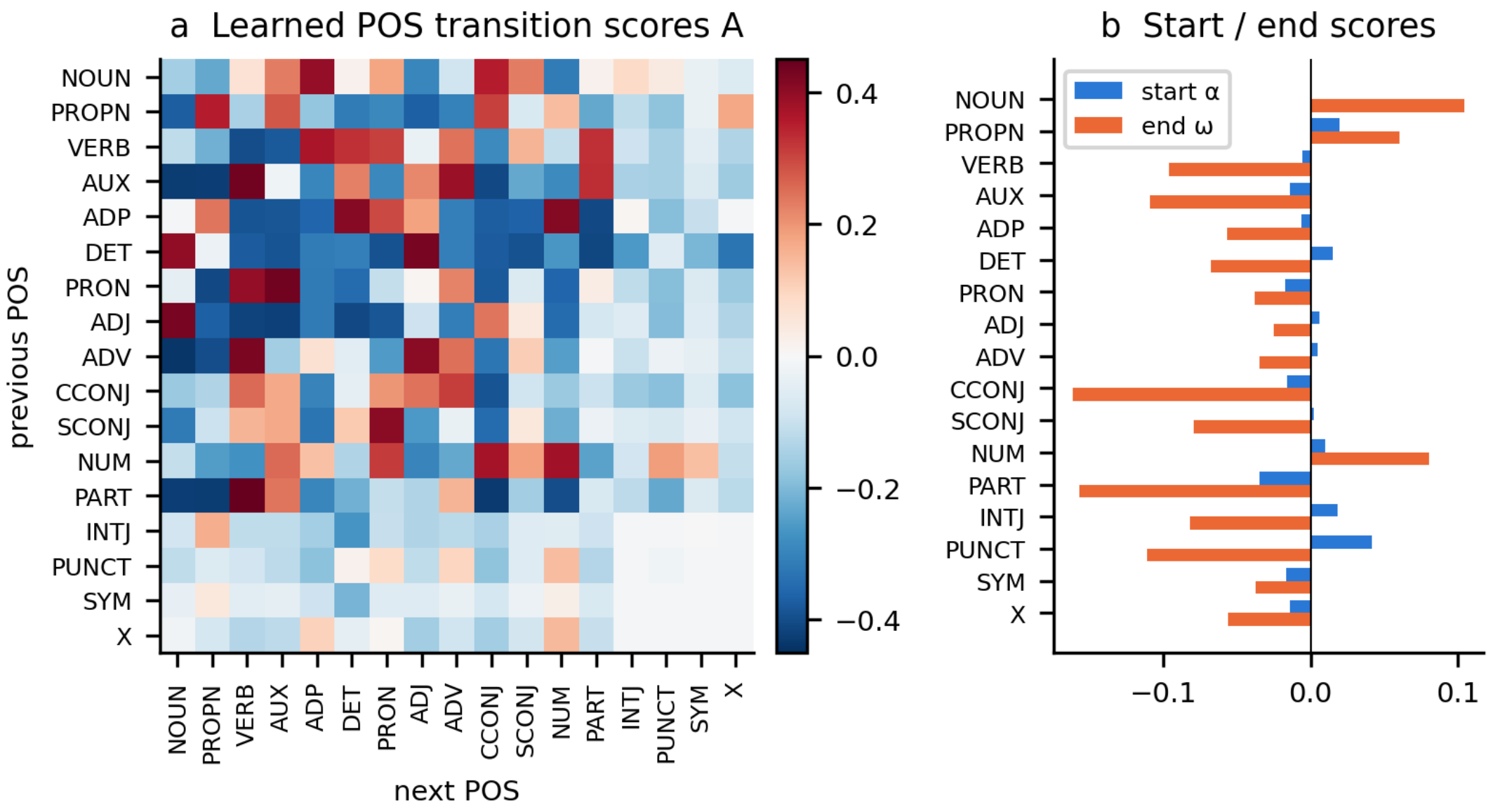

**Fig. 11.** Learned parameters of the joint word–POS CRF. (a) Transition scores $A$ between Universal POS tags. (b) Start and end scores.

These transitions encode the internal order of English noun and verb phrases: determiners and adjectives precede nouns, auxiliaries and the infinitive marker precede verbs, and nouns rarely follow adverbs or auxiliaries directly. Because the lattice contains only fixated words, the CRF learned these regularities over sequences from which skipped words, mostly short function words, had been removed; that phrase-level order survives in these gapped sequences shows that the fixated words retain much of the syntactic skeleton of the sentence. The transitions are estimated from the training sentences not from the EEG. They supply grammatical structure that single-fixation EEG cannot provide (single-word class AUC 0.53; Section 3.3) and turn weak, graded POS evidence into well-formed tag sequences. Unlike the syntactic knowledge implicit in a language model, this structure is a 17 × 17 matrix that can be read, checked and constrained, which makes the grammatical decisions of the decoder inspectable.

#### 5.3.2. Semantic and POS head outputs in a single recording

Figure 12 shows one test recording selected from 87 candidates using EEG and eye-movement criteria: the share of words with a clear fixation response, baseline noise, and longer gaze on content than on function words. The reader fixated “Whiffle-Ball” six times for 1.4 s and skipped “II”, “–” and “as”. The single-word EEG traces fluctuate by about $\pm 10\ \mu$V around a reader-average response of about 1 $\mu$V, which illustrates the signal-to-noise problem at the level of one word. The receptive field of the trained encoder is concentrated on and around each word, whereas attention reaches distant words. Figure 13 follows the same recording through the decoder. The top candidates at each position are words typical of movie reviews (“unwatchable”, “shapeless”, “unimaginative”), each only 1.3–1.6 times more probable than a uniform choice among 23,890 words; the gold words ranked between 30th and 7,301st where they were in the lexicon. The output combines review-style EEG-position words with inserted function words and punctuation.

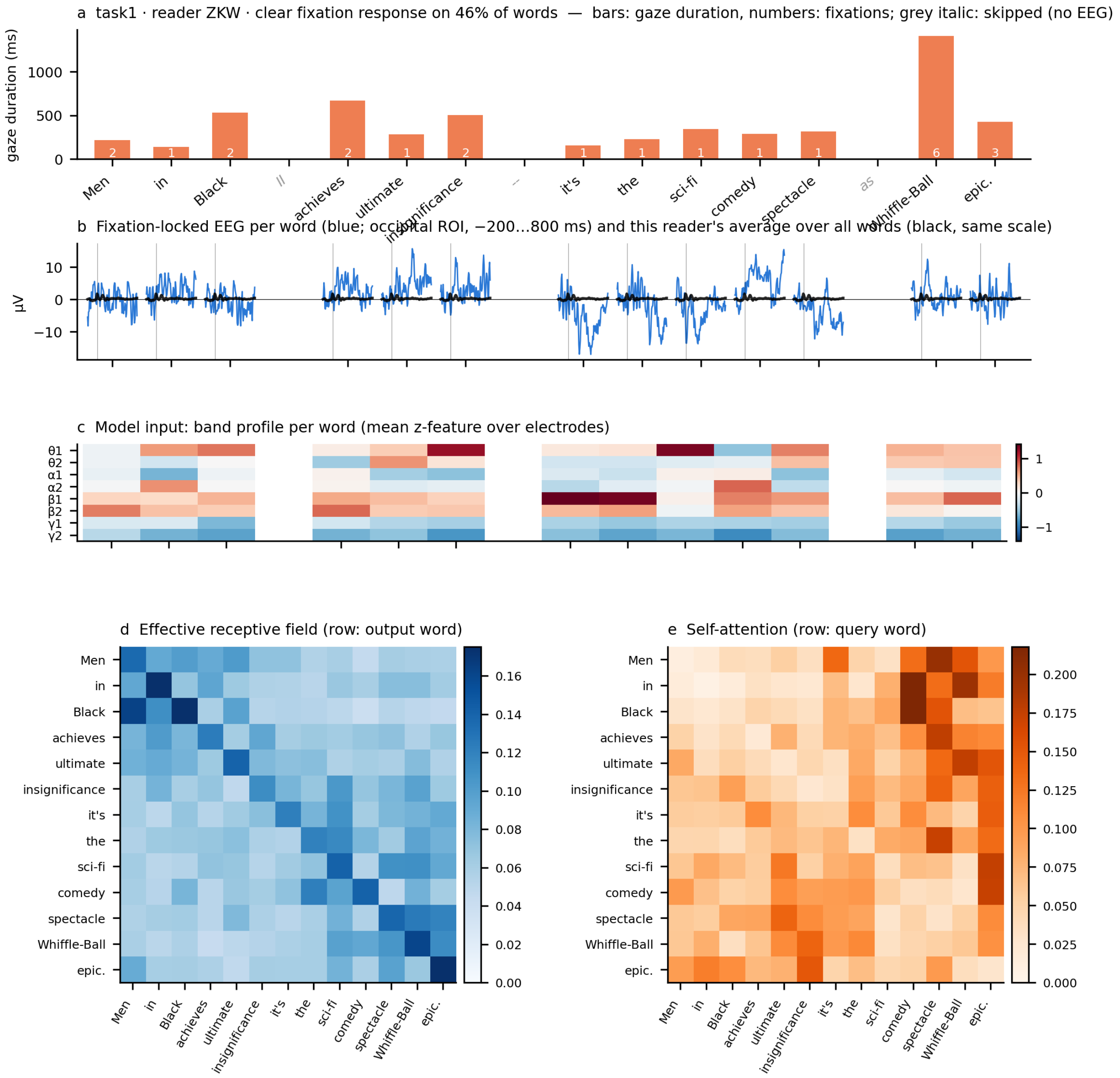

**Fig. 12.** One test recording (reader ZKW, SR task). (a) Gaze duration and number of fixations per word; skipped words are shown in grey. (b) Fixation-locked EEG per EEG word at the occipital ROI, displayed as a word-by-time heat map. (c) Mean band profile of the model input. (d) Effective receptive field of the trained encoder. (e) Self-attention weights.

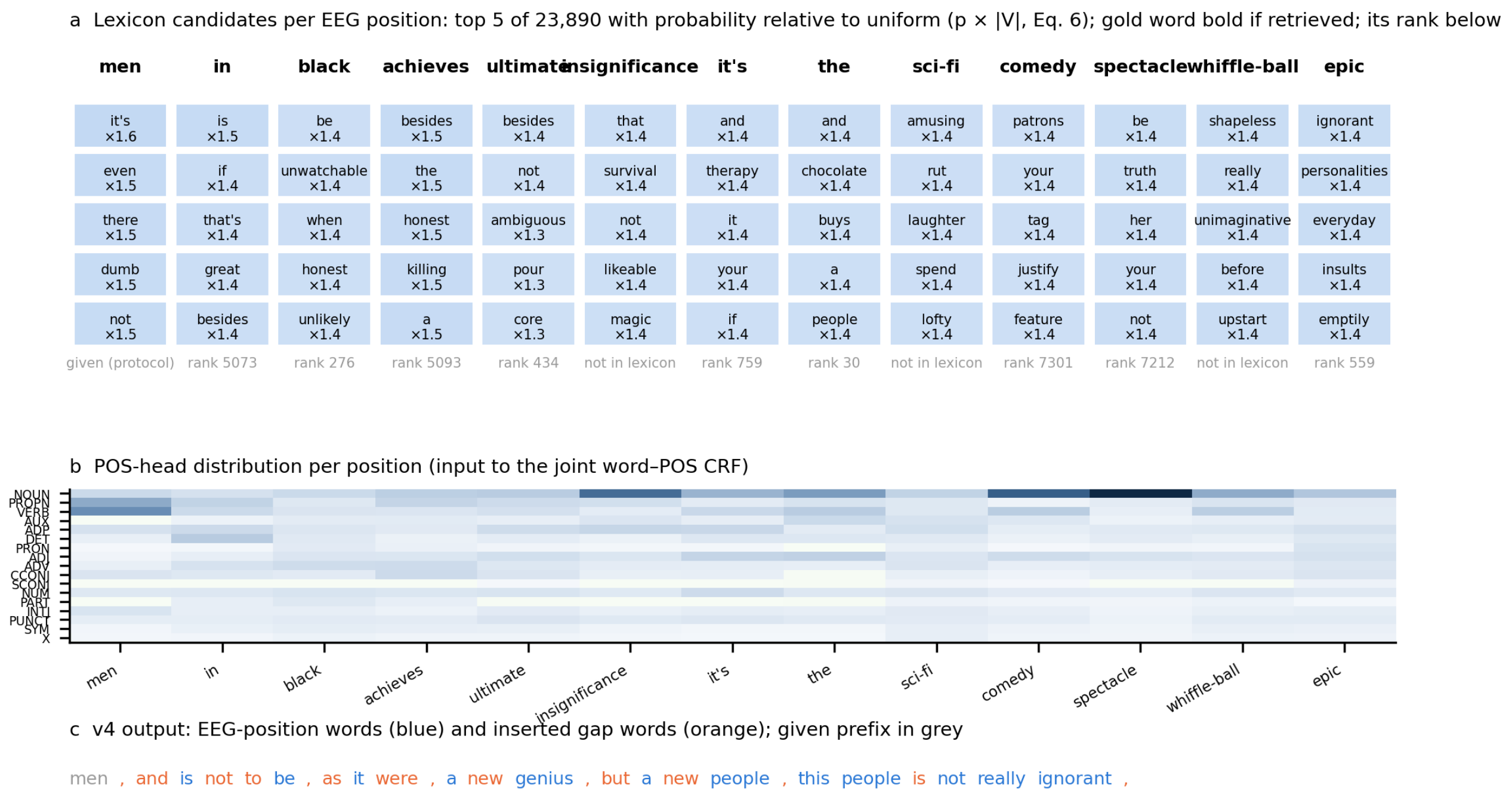

**Fig. 13.** Decoding of the recording in Fig. 12. (a) Top five lexicon candidates per EEG position with their probability relative to uniform; the rank of the gold word is shown below. (b) POS-head distribution. (c) Output of EEGHyena2 with EEG-position words in blue, inserted gap words in orange, and the reference below.

This recording shows what the semantic head extracts from one fixation. Its top candidates matched the genre of the text, evaluative words typical of movie reviews, but not the words that were read, and its probabilities were close to uniform. This pattern is expected if word-level EEG mainly carries information shared by many words of a recording, such as the reader, the task and the type of text, and little that is specific to one word, which agrees with the strong reader identification and weak single-word classification in Section 3.3. All sentiment-reading sentences in ZuCo are movie reviews, so features that identify the reader and the task also predict the vocabulary of the task. A decoder that committed to the top word would therefore produce on-topic but wrong words. Keeping a broad candidate set and passing calibrated uncertainty to the search, as advocated for neural decoding of EEG more generally (Duan et al., 2023), leaves the choice among them to the CRF and the language model. The output is consistent with this account: it reads as a plausible review with correct function-word scaffolding but shares few content words with the reference, which places the main bottleneck in the lexical evidence rather than in the search.

### 5.3.3. Backbone gates and encoder

The three channel–frequency gates stayed near pass-through: after training, all weights lay between 0.980 and 0.985 (initial value 0.982). The trained encoder used all three ranges and mixed information from both sides of each word (Fig. 14). The short, medium and long operators placed 70%, 43% and 27% of their kernel weight within two words. Attention down-weighted the word itself (0.4 times uniform) and its immediate neighbours and favoured words three to eight positions away, so the convolutions supplied local context and attention distant context. The effective receptive field peaked at the word itself (about 15% of the input gradient) and decayed nearly symmetrically with distance.

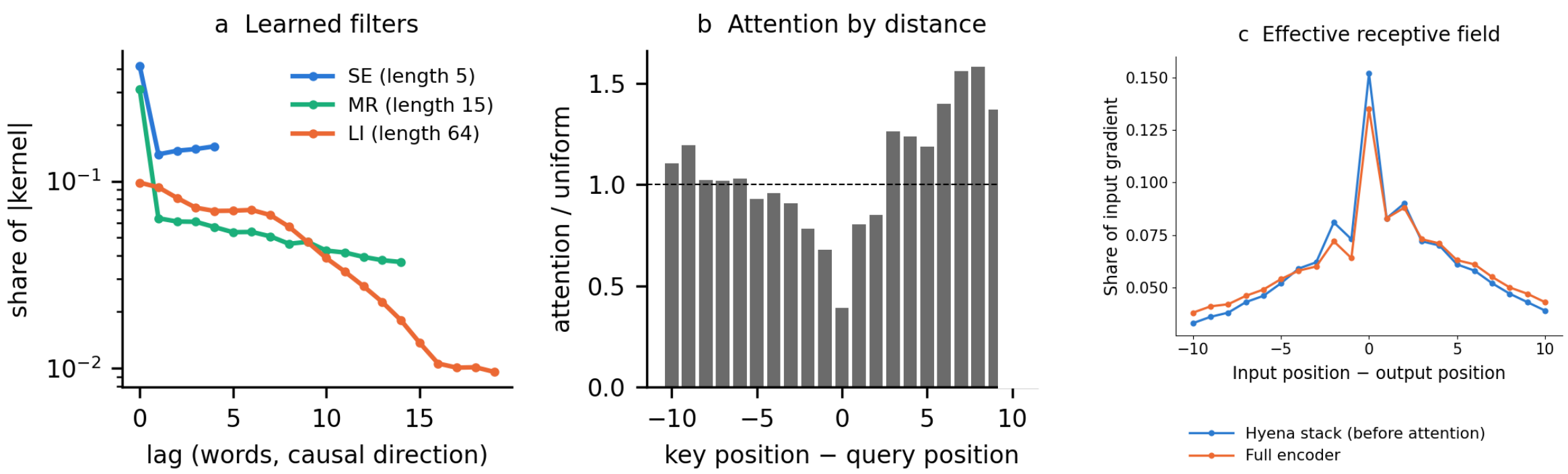

**Fig. 14.** What the trained Bi-SH2 encoder uses (291 eligible test sentences from 300 selected). (a) Share of absolute kernel weight by lag for the short (SE), medium (MR) and long (LI) operators. (b) Self-attention weight by signed word distance relative to uniform attention. (c) Effective receptive field (Eq. 17) for the Hyena stack and the full encoder.

The near-constant gates are a null result for spatial–spectral selection: no band or electrode was up- or down-weighted, consistent with the widespread but weak class differences in Fig. 4. This should not be taken as evidence that all features are equally informative, because the gates were initialised in the saturated range of the logistic function, where gradients are small ($\sigma'(4) \approx 0.018$), and may have changed little whatever the information in the features. The encoder, in contrast, developed a clear division of labour, with the convolutions concentrated on nearby words and attention on words three to eight positions away, so that the two operator types complemented rather than duplicated each other. The local weighting matches the timing of reading, in which 71% of next-word fixation onsets fell inside the current word's epoch, yet Section 3.3 found no word-specific information in the band-power features of neighbouring words. A more plausible role of the local mixing is noise reduction: pooling the representations of neighbouring fixations raises the signal-to-noise ratio of information shared across words, as averaging raised the class AUC in Fig. 5, while word-specific information remains limited. The symmetric receptive field also shows that the encoder drew on following words as much as on preceding ones, which is possible because decoding takes place after the whole sentence has been read.

# 6. Conclusion

We introduced EEGHyena2, a structured EEG-to-text decoder in which a bidirectional Hyena-style encoder feeds a semantic head and a POS head, and a joint word–POS CRF turns their graded lexical and grammatical evidence into a restricted lattice that a frozen Llama-2 scores in context. The design followed from fixation-related analyses showing that word class is expressed reliably in group-averaged EEG but only weakly in single words. Tested on ZuCo and ROAMM, the decoder matched the closest published method on BLEU-1, improved ROUGE-1 F1, decoded several times faster than LLM candidate search and exceeded length-matched Llama-2 continuations in ROUGE-L F1; NeuroLM-B remained faster, with metric-dependent trade-offs.

The advantage over length-matched language-model-only continuations combines the word–POS interface, the search policy and any input effects, and replacing the EEG-derived emissions with training-set averages lowered overlap at a similar output length. These results suggest that weak word-level EEG evidence is best used when it is pooled across the sentence through explicit lexical and grammatical structure, and that language models can serve brain-to-text decoding as adjudicators rather than generators, which lowers decoding cost and keeps each decision inspectable. Recovery of word content from the EEG nonetheless remains limited: beyond the given first word, every method recovered fewer than 3% of the reference content words.

Future work will extend EEGHyena2 to larger and more linguistically diverse natural-reading datasets, with broader candidate vocabularies, more independent participants and texts, and evaluation under reduced reference-prefix constraints. A central direction is to identify how EEG-derived information, structured decoding constraints, and language-model priors contribute to reconstruction. The train-marginal semantic/POS replacement presented here is an initial within-decoder control; in the future, factorial replacements of the two emission channels and distribution-matched counterfactuals are needed for finer causal attribution. Complementary ablations of candidate vocabulary, POS/CRF transitions, skipped-word insertion and length constraints, with matched realised output lengths, would clarify how these components interact. Evaluating both intermediate candidate rankings and final generated content could identify where useful EEG-dependent information is preserved or lost.

Another important question concerns the information captured by pretrained EEG representations and its overlap with linguistic knowledge already available to the LLM. Comparing frozen and adapted EEG encoders within the same decoding framework, with subject-specific calibration and held-out-participant evaluation, could reveal whether neural pretraining improves stimulus-specific, lexical, syntactic, or semantic information. Measuring the incremental predictive value of EEG after controlling for LLM probabilities, word frequency, lexical form, and eye-movement features would further distinguish complementary neural information from shared language and behavioral priors.

Finally, fixation-aligned temporal EEG representations and richer eye-tracking integration (Zhu et al., 2024) may offer more direct access to the overlapping neural processes of natural reading than aggregated band-power features alone. Earlier exploratory analyses suggest that EEG–language relationships may vary across reading tasks and participants, motivating comparisons across cognitive conditions and levels of linguistic representation. These experiments would identify when neural signals contribute information beyond language-model expectations.

# Declaration of competing interest

None.

# Data and code availability

The ZuCo 1.0 and 2.0 datasets are publicly available from the Open Science Framework (Hollenstein et al., 2018, 2020), and ROAMM is available from OpenNeuro as ds007629 (Sun et al., 2026). The code is available from the corresponding author upon reasonable request.

# Declaration of generative AI and AI-assisted technologies in the manuscript prepration process

During preparation of this manuscript, the authors used OpenAI Codex and the Scientific Agent Skills library (Kassis et al., 2026) to support English-language editing, structural review and reference checking. After using this tool/service, the authors reviewed and edited the content as needed and take full responsibility for the content of the published article.

# Acknowledgement

The authors acknowledge the use of the Apollo GPU cluster, a service provided by the Sydney Informatics Hub, a Core Research Facility of the University of Sydney.